\documentclass[11pt,letterpaper]{article}
\usepackage{newtxtext}
\usepackage[margin=1in]{geometry}
\usepackage{amsmath,amssymb}
\usepackage{microtype}
\usepackage[dvipsnames]{xcolor}
\usepackage{graphicx}
\usepackage[export]{adjustbox}
\graphicspath{{./}}
\let\originalincludegraphics\includegraphics
\renewcommand{\includegraphics}[2][]{\originalincludegraphics[#1,max width=\linewidth]{#2}}
\usepackage{booktabs,multirow,tabularx}

\usepackage[font=small,labelfont=bf,labelsep=period,skip=8pt]{caption}
\usepackage{float}
\usepackage{physics}
\usepackage{seqsplit}
\usepackage[numbers,sort&compress]{natbib}
\usepackage{xurl}
\usepackage{hyperref}
\hypersetup{colorlinks=true, urlcolor=magenta}
\hypersetup{pdftitle={Highly Deformable Proprioceptive Membrane for Real-Time 3D Shape Reconstruction},pdfauthor={Guanyu Xu, Jiaqi Wang, Dezhong Tong, Xiaonan Huang}}
\begin{document}
\begin{center}
{\LARGE\bfseries Highly Deformable Proprioceptive Membrane for Real-Time 3D Shape Reconstruction\par}
\medskip
{\large Guanyu Xu$^{\dagger}$, Jiaqi Wang$^{\dagger}$, Dezhong Tong, Xiaonan Huang$^{*}$\par}
\medskip
{\small Robotics Department, University of Michigan, Ann Arbor, MI, USA\\
Email Address: xuguanyu@umich.edu, wangjq@umich.edu, dezhong@umich.edu, xiaonanh@umich.edu\\
$^{*}$Corresponding author. $^{\dagger}$These authors contributed equally to this work.\par}
\end{center}

\begin{abstract}
Reconstructing the three-dimensional (3D) geometry of surfaces is essential for robot perception and physical interaction. Vision-based approaches reconstruct surfaces at high resolution from an external viewpoint but often require a clear line of sight. A conformable membrane that reconstructs its own deformation offers a complementary route. Based on optical waveguide shape sensing, this work presents a stretchable proprioceptive silicone membrane that reconstructs localized contact and in-plane stretching under large deformation. The membrane integrates edge-mounted light-emitting diodes (LEDs) and centrally distributed photodiodes (PDs) in an elastomeric waveguide using liquid-metal interconnects. A neural network maps the optical intensity measurements to a dense point cloud of the membrane surface. Optical acquisition and model inference run in real time at about 90~Hz. Under localized pressing of up to about 25~mm depth, the membrane achieves an overall shape reconstruction error of 1.580~mm. It also localizes finger contacts with a mean error of 1.779~mm. Extreme strain tests show that the optical signals remain informative for in-plane outline reconstruction up to 75\% strain, approaching the mechanical limit of the liquid-metal circuit. These results indicate a practical route to proprioceptive shape sensing in deformable robotic systems. Videos can be found at \href{https://proprioceptive-membrane.github.io/}%
     {\textcolor{magenta}{{https://proprioceptive-membrane.github.io/}}}.
\end{abstract}

\section{Introduction} \label{sec:intro}
Perceiving the geometry of large-scale three-dimensional (3D) surfaces is a fundamental capability in robotics, as it provides rich information for both physical-interaction reasoning and state estimation~\cite{zhang2025artificial}. Reconstructing the surface geometry of external objects enables tasks such as object pose estimation~\cite{suresh2024neuralfeels, murali2025shared} and grasp planning~\cite{varley2017shape, han2025morphing}.
Integrating shape-sensing capabilities into soft robotic actuators provides proprioceptive feedback for state estimation~\cite{fan2025sparc, chen2026sppiro} and closed-loop control~\cite{wang2024sensing, micklem2026harnessing}. In addition, shape sensing can provide direct information about the properties of objects in contact~\cite{yu2025recent,yuan2017gelsight}, as well as enable inference of force distribution or contact locations~\cite{fang2025force, park2022biomimetic}.

Vision-based methods, including multi-view stereo~\cite{schonberger2016pixelwise,wang2026learning}, depth-camera fusion~\cite{newcombe2011kinectfusion, zollhofer2018state}, and neural scene representations~\cite{mildenhall2021nerf, kerbl20233d}, are widely used in 3D surface reconstruction and recover detailed geometry without physical contact. Because these methods observe a surface from a distance, their performance depends on illumination and on a clear line of sight~\cite{smith20203d}. In manipulation and soft robotics, the surface of interest is often the contact interface itself, where suitable viewing conditions are difficult to guarantee. A thin, conformable membrane that adheres to the surface and reconstructs its own deformation provides complementary geometric information that is independent of an external viewpoint. These considerations motivate \textit{proprioceptive shape sensing}, in which a thin membrane is sensorized to reconstruct its deformation directly from internal measurements.

\begin{figure}[!htb]
    \centering
    \includegraphics{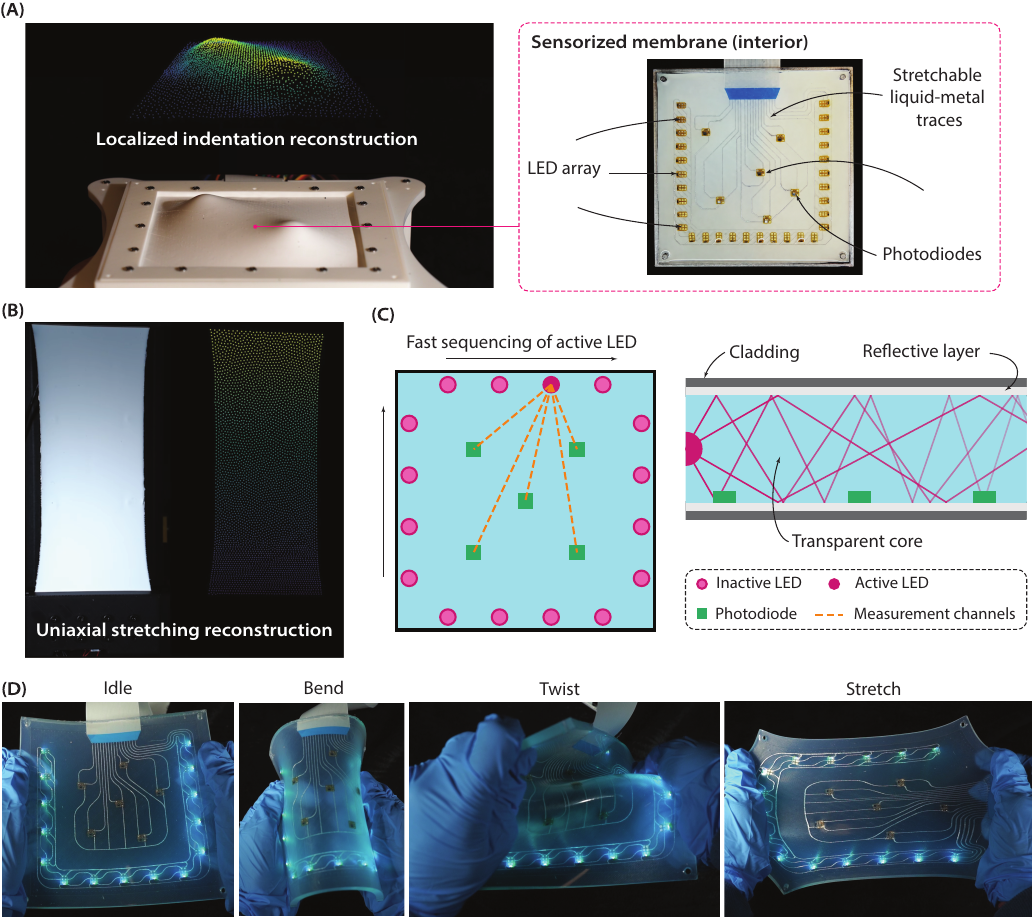}
    \caption{\textbf{Stretchable optical waveguide membrane for proprioceptive shape reconstruction.} \textbf{(A)} Localized indentation reconstruction on the 140~mm $\times$ 140~mm membrane, and the interior of the sensorized membrane with the edge LED array, the central photodiodes, and the stretchable liquid-metal traces. \textbf{(B)} In-plane reconstruction of the rectangular prototype under uniaxial stretching. \textbf{(C)} Time-multiplexed readout and sensing principle: the LEDs are switched on one at a time, and all photodiodes are sampled while each LED is on, giving a measurement matrix per frame (schematic). The cross section shows the transparent core between the reflective and cladding layers. \textbf{(D)} Photographs of a membrane without encapsulation layers under bending, twisting, and stretching. Reconstruction of bending and twisting is not evaluated in this work.}
    \label{fig:teaser}
\end{figure}

Researchers have explored multiple methods to sensorize thin membranes for proprioceptive 3D shape reconstruction. Early work~\cite{mittendorfer20123d, hoshi20083d} assembled multiple rigid modules into flexible, mesh-like structures, with each module carrying an inertial measurement unit (IMU) to estimate local orientation. Soft sensing skins subsequently adopted resistive~\cite{park2012design, truby2020distri}, capacitive~\cite{hu2024touch}, and magnetic~\cite{yan2021soft} mechanisms, which measure the spatial distribution of local strain or field changes and recover the global geometry from these distributed measurements~\cite{shu2023machine}. Shah et al. combined discrete IMUs and capacitive strain sensors in a single stretchable shape-sensing (S3) sheet~\cite{shah2023stretchable}. In these designs, as the number of electrical sensing elements grows, wiring, signal cross-talk, and electromagnetic interference (EMI) become central design considerations~\cite{shih2020electronic}. Optical-based sensing offers a complementary route~\cite{wang2024soft}.

Optical-based sensors encode deformation either in a characteristic wavelength or in a transmitted light intensity. In wavelength-based sensing, axial strain modulates the reflected wavelength of fiber Bragg gratings (FBGs), enabling high-resolution deformation measurements~\cite{FLORIS2021106508}. FBGs embedded in silicone sheets have enabled real-time reconstruction of surface shape over large areas~\cite{lun2019real,wang2021large}, and FBG sensing has also supported wearable motion tracking~\cite{feng2023optical,hu2023fiber} and large-area tactile skins~\cite{massari2022functional}. These systems are read out with optical interrogators, and the embedded fibers largely determine the composite mechanics, making them well suited to flexible structures dominated by bending but less compliant under large strains.

Intensity-based optical waveguide sensing, on the other hand, relies on light-emitting diodes (LEDs) and photodiodes (PDs) integrated into soft structures, enabling multi-degree-of-freedom motion estimation in soft actuators, prosthetic hands, and human--robot interfaces~\cite{krauss2024enhanced, kim2020heterogeneous, zhao2016optoelectronically, heiden20223d, krauss2022stretchable}. Later designs detect deformation along a stretchable waveguide~\cite{bai2020stretchable}, resolve the bending direction within a single waveguide~\cite{lee2025multimodal,julia2026enabling}, and extend sensing to lattices and sheets built from multiple waveguides~\cite{yu2026toward}. Mak et al. pioneered the extension of intensity-based sensing from one-dimensional waveguides to a continuous two-dimensional waveguide for surface shape reconstruction~\cite{mak2024intelligent}. In their work, three pairs of RGB LEDs and color photodetectors allow an autoregressive network to reconstruct 3D shape at a sparse set of surface nodes during combined bending and twisting, while local indentation and in-plane stretching remain an open challenge.

To address this challenge, we explore optical waveguide sensing in a \textit{stretchable} membrane. In this work, we present a highly deformable proprioceptive membrane for real-time 3D shape reconstruction using intensity-based optical sensing. We develop a layer-by-layer fabrication method that embeds the optoelectronic components and a stretchable circuit within a cast multilayer elastomeric waveguide. In this method, all LEDs and PDs are connected through liquid-metal traces laser-patterned on a stretchable substrate, so the circuit deforms with the elastomeric composite (Figure~\ref{fig:teaser}D). We place LEDs along the membrane edge and distribute the PDs over the central region (Figure~\ref{fig:teaser}A). By sequentially activating the LEDs and recording the corresponding light intensities at the PDs, we obtain a rich set of deformation-dependent signals (Figure~\ref{fig:teaser}C). We further develop a neural network that maps each signal set to a dense point cloud of the entire membrane surface.

We characterize the sensor in multiple deformation regimes, including bending and stretching. We achieve high-accuracy surface shape reconstruction for localized indentation (Figure~\ref{fig:teaser}A) and demonstrate in-plane reconstruction under uniaxial stretching (Figure~\ref{fig:teaser}B). For localized indentation, a 140~mm square membrane clamped on all four sides is pressed with indenters of different shapes. The reconstructed surface reaches a mean reconstruction error of 1.580~mm for deformation magnitudes of up to about 25~mm. Along continuous finger strokes held out from training, the reconstruction localizes the contact with a mean error of 1.779~mm. In an extreme strain test, we stretch a 120~mm $\times$ 100~mm rectangular prototype on a tensile tester from 0\% to 75\% strain, near the mechanical limit of the liquid-metal circuit, and show that the optical signals still encode shape information for in-plane outline reconstruction. The system reconstructs the membrane shape end-to-end in real time at about 90~Hz, from optical acquisition to network inference on a laptop. A grouped Shapley additive global importance (SAGE) analysis further quantifies how individual LEDs and PDs contribute to the reconstruction, providing guidance for sensor layouts.  The real-time reconstruction of localized indentation and in-plane stretching suggests that the membrane could be deployed in deformable robotic systems for state estimation and tactile sensing, particularly where external cameras are occluded by contact or limited in viewpoint.
\section{Design and Fabrication of Membrane Sensor} \label{sec:hardware-design}
The membrane sensor measures deformation-dependent changes in light intensity using LEDs and PDs embedded in a stretchable optical waveguide. This section describes its mechanical structure, electrical connections, readout scheme, and fabrication process.

\begin{figure}[!htbp]
    \centering
    \includegraphics[width=\linewidth]{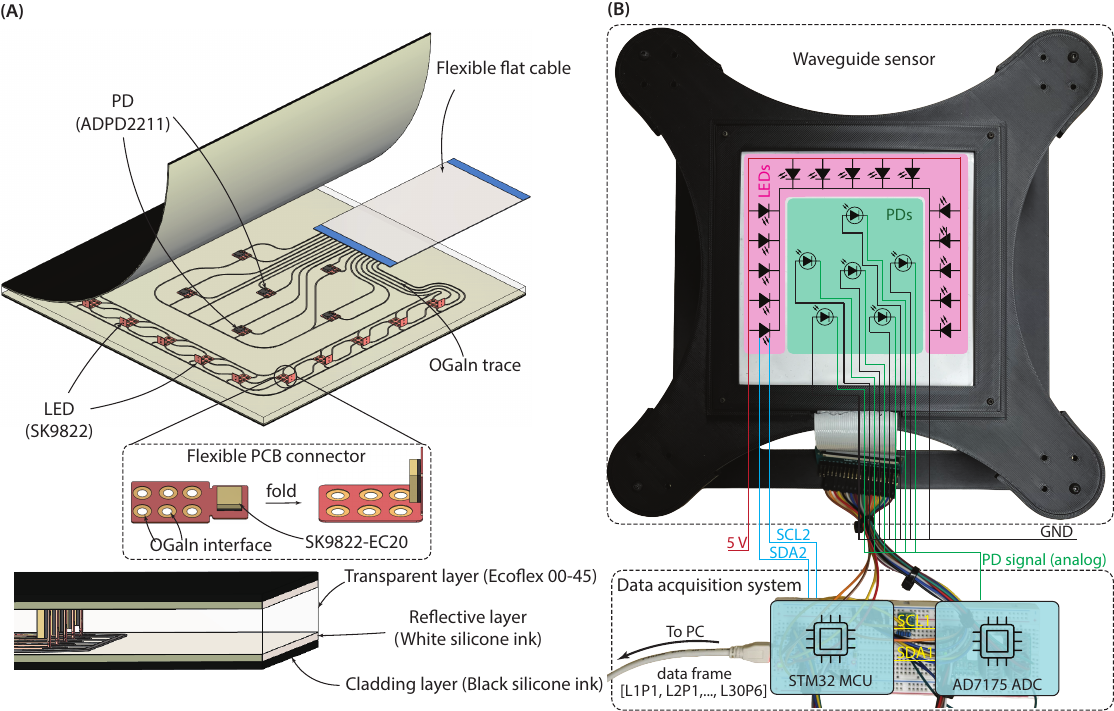}
    \caption{\textbf{Hardware architecture and data acquisition system.} \textbf{(A)} Exploded CAD view of the optical waveguide sensor showing the internal layer stack, photodiode and LED array, stretchable OGaIn interconnects, and flexible flat cable interface. Flexible PCB connectors are designed to fold upward to mount LEDs vertically in the waveguide. \textbf{(B)} Photograph of the sensor prototype with the electrical wiring overlaid. An STM32 microcontroller sequences the LEDs and coordinates photodiode sampling via the external ADC, producing a complete measurement frame by concatenating responses across all LED--photodiode pairs.}
    \label{fig:hardware}
\end{figure}

\subsection{Mechanical System Design} \label{sec:mechanical-system}
The waveguide structure confines light propagation within its transparent core while reducing sensitivity to external illumination. As shown in Figure~\ref{fig:hardware}A, the device adopts a five-layer stack. A transparent elastomer core is sandwiched between two reflective (white) layers, which are in turn enclosed by two outer cladding (black) layers. The reflective layers scatter light back into the core and reduce optical loss into the cladding, whereas the black cladding attenuates incident ambient light to reduce its influence on the optical measurements. The core material is selected for high optical transmittance while remaining compliant under large deformations.

Light-emitting diodes (LEDs) and photodiodes (PDs) are embedded within the core and differ in their mounting orientation (Figure~\ref{fig:hardware}A). The LEDs are mounted upright along the membrane edge on flexible printed circuit board (PCB) breakouts folded by 90$^\circ$, with their emitting faces directed inward so light is injected into the core plane. This edge injection directs the emitted light along the core toward the PDs across the membrane. In contrast, the PDs lie flat with their sensitive faces parallel to the membrane surface and are distributed over the central region to intercept light from many directions. Both component types are carried on a single stretchable substrate offset from the core mid-plane toward one face of the membrane.

The in-plane layout of LEDs and PDs is designed to capture complementary deformation-dependent optical responses. LED--PD pairs with different in-plane orientations can provide complementary sensitivity to deformation along different directions (Section~\ref{sec:response}). In practice, two constraints guide this layout. Light intensity attenuates strongly with propagation distance inside the waveguide, and the PD cannot resolve arbitrarily small signals, so an LED--PD pair remains informative only over a bounded distance range. In the present stack, a PD reading saturates the analog-to-digital converter (ADC) at separations below about 29~mm and approaches the noise floor beyond about 80~mm (Figure~\ref{fig:SI-pdreadings}).
In addition, every signal line is routed through the stretchable circuit and the connector to the acquisition board, and wiring complexity is a constraint when the number of sensing elements in a soft skin grows~\cite{shih2020electronic}. Each PD requires its own signal line, whereas all LEDs share a common bus (Section~\ref{sec:electrical-system}), so the wiring limits the number of PDs much more strongly than the number of LEDs.
These optical and wiring constraints jointly limit the membrane area a single sensing unit can effectively cover. The resulting layout combines peripheral LEDs with PDs distributed across the central region to provide multiple pair orientations within a compact sensing area.

\subsection{Electrical System Design} \label{sec:electrical-system}
To maintain stretchability, we use oxidized gallium--indium alloy (OGaIn) \cite{woodman2024stretchable, bartlett2016rapid} patterned on a stretchable substrate to connect all electrical components. The dispersed oxide gives OGaIn a paste-like consistency that holds a painted pattern, which simplifies the fabrication process. Each bare optoelectronic component interfaces with the OGaIn traces via a small flexible PCB breakout, which provides strain relief and a mechanically robust electrical connection. These breakouts and components introduce localized stiffening in their immediate vicinity, while the liquid-metal traces remain compliant and stretchable.

Sensor readout uses sequential illumination and synchronous readout (Figure~\ref{fig:flowchart}). At each illumination step, the system activates a single LED and samples all photodiode channels simultaneously. The illumination interval accommodates signal settling and acquisition while limiting the total scan time. The procedure then advances to the next LED until it scans the entire LED array, yielding one measurement for each LED--PD pair. For a membrane with $\ell$ LEDs and $p$ PDs, these measurements are concatenated into a single high-dimensional measurement vector $\boldsymbol{v}=[v_{1,1},\dots,v_{\ell,1}, \dots, v_{1,p}, \dots, v_{\ell,p}]^{\top}\in\mathbb{R}^{p\ell}$, where $v_{i,j}$ is the reading of PD $j$ while LED $i$ is on. Sequential illumination separates the contributions of individual LEDs, allowing the reconstruction model to use the combined responses of multiple LED--PD pairs.
Because the number of measurement channels grows with the product $p\ell$ and the number of interconnects grows only with $p$, we use many more LEDs than PDs to increase measurements while limiting wiring complexity. Each additional LED instead lengthens the scan, so $\ell$ is balanced against the frame rate. With 30 LEDs and 6 PDs, each frame contains 180 LED--PD measurements, and a complete scan is acquired about every 11~ms, which corresponds to about 90~Hz.

\begin{figure}[!ht]
    \centering
    \includegraphics[width=\linewidth]{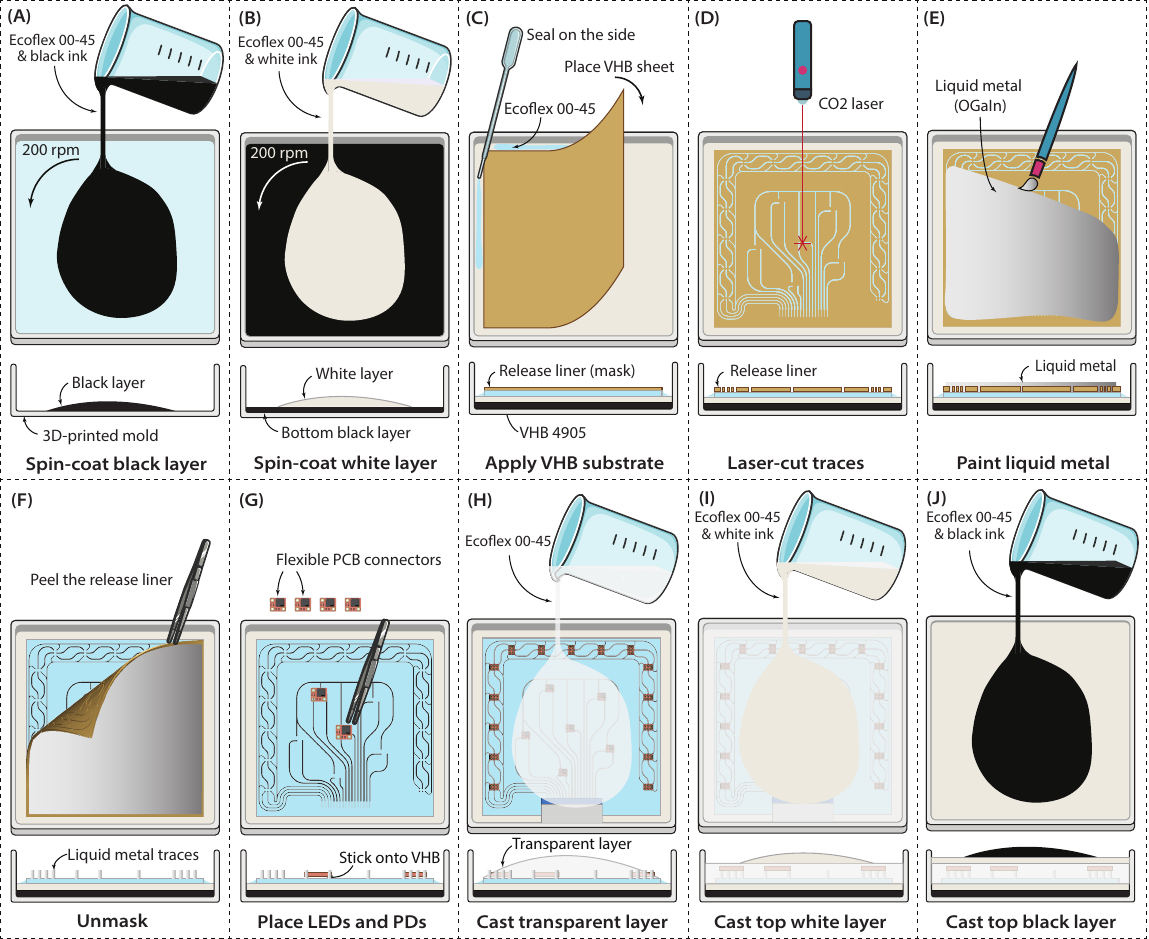}
    \caption{\textbf{Manufacturing process of the optical waveguide sensor.} \textbf{(A)} Spin-coat the bottom cladding (black) layer in a 3D-printed mold. \textbf{(B)} Spin-coat the bottom reflective (white) layer on top of the black layer. \textbf{(C)} Place a piece of VHB 4905 sheet on the cured white layer and seal the side with elastomer (Ecoflex 00-45). \textbf{(D)} Cut circuit traces on the  release liner of the VHB using a laser cutter. \textbf{(E)} Paint liquid metal (OGaIn) on the  release liner to fill the traces. \textbf{(F)} Peel the  release liner and leave the OGaIn traces on the VHB. \textbf{(G)} Attach the flexible PCB breakouts carrying the LEDs and PDs to the traces. \textbf{(H)} Cast the transparent core of the waveguide. \textbf{(I)} Cast the top reflective layer. \textbf{(J)} Cast the top cladding layer, and demold the entire sensor after it is fully cured.}
    \label{fig:fabrication}
\end{figure}

Figure~\ref{fig:hardware}B shows the electrical schematic. The electrical architecture supports rapid multiplexed acquisition. Addressable LEDs connected in a daisy chain share a common serial interface while remaining individually controllable, enabling sequential illumination with few interconnects. A high-resolution ADC digitizes analog photodiode outputs, capturing deformation-dependent intensity variations across channels with different signal levels. A microcontroller coordinates illumination and sampling to associate each measurement with the active LED and transfers the acquired signals to the host computer running the reconstruction model. Table~\ref{tab:SI-components} lists the components and interface specifications. We switch off all LEDs once per frame and subtract the resulting no-light reading of each PD from that frame's readings before transmitting the measurements to the host  computer. This subtraction accounts for potential signal offsets that do not originate from the LEDs.

\subsection{Sensor Fabrication} \label{sec:fabrication}
The sensor fabrication follows a layer-by-layer stacking process that embeds the liquid-metal circuit and optoelectronic components inside a cast silicone waveguide. Each layer is built directly on the cured bottom layers inside the casting mold, and the entire stack is demolded  in the end. Figure~\ref{fig:fabrication} illustrates the process. The bottom black cladding and white reflective layers are sequentially cast and cured in a mold (Figure~\ref{fig:fabrication}A--B). An adhesive VHB sheet placed on the reflective layer serves as the circuit substrate, and its release liner serves as the patterning stencil. The trace layout is laser-cut into the liner in place, and the cut regions are peeled off to expose the adhesive (Figure~\ref{fig:fabrication}C--D). OGaIn painted over the liner adheres to the exposed adhesive, and removing the remaining liner leaves the patterned traces on the substrate (Figure~\ref{fig:fabrication}E--F).

The silicone layers of the stack bond directly to one another during sequential casting, whereas silicone does not bond to the VHB substrate. We therefore cut the substrate smaller than the membrane footprint and fill the resulting perimeter margin with silicone elastomer. This seal later joins the bottom reflective layer to the transparent core, fully enclosing the VHB substrate in bonded silicone and preventing the membrane from delaminating from the substrate. After verifying electrical continuity, we attach the flexible flat cable and the optoelectronic components to the patterned substrate, with the LED breakouts folded upright (Figure~\ref{fig:fabrication}G). The adhesive substrate provides mechanical fixation, and the liquid metal forms the electrical contact at the breakout pads. We then cast the transparent core around the integrated components, which embeds the LEDs and PDs in the core (Figure~\ref{fig:fabrication}H), followed by the top reflective and cladding layers with the same formulations as the bottom layers (Figure~\ref{fig:fabrication}I--J). A single laser-cut pattern defines the trace layout, so changing the number or position of the LEDs and PDs requires only a new cutting pattern. Section~\ref{sec:SI-fabrication-detail} provides details of material formulations, fabrication dimensions, and processing conditions.

\section{Neural Network for Shape Reconstruction} \label{sec:neural-network}
\begin{figure}[!htbp]
    \centering
    \includegraphics{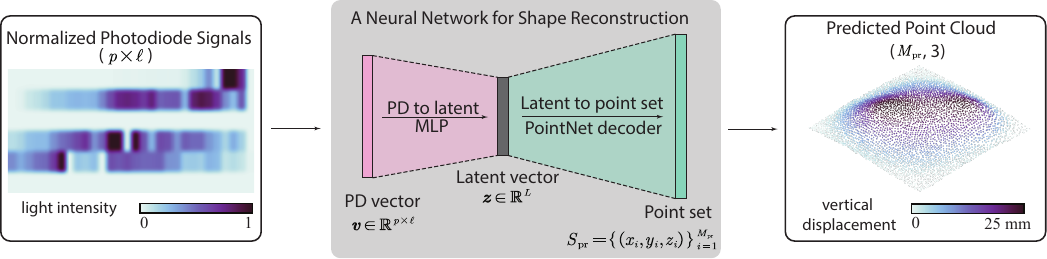}
    \caption{\textbf{Two-stage shape reconstruction model.} The normalized photodiode signals of one LED scan ($p\times\ell$) are flattened into the measurement vector $\boldsymbol{v}\in\mathbb{R}^{p\ell}$. A PD-to-latent MLP maps $\boldsymbol{v}$ to a latent vector $\boldsymbol{z}\in\mathbb{R}^L$, and the decoder of a pre-trained point-cloud autoencoder maps $\boldsymbol{z}$ to the predicted point set $S_{\mathrm{pr}}$ with $M_{\mathrm{pr}}$ points, colored here by vertical displacement. Figure~\ref{fig:SI-model} details both training stages.}
    \label{fig:model}
\end{figure}

To accurately reconstruct the sensor's 3D geometry from optical signals, we implement a two-stage data-driven framework (Figure~\ref{fig:model}). The approach first establishes a low-dimensional latent space of feasible sensor deformations using a point-cloud autoencoder, effectively embedding geometric constraints into the model. Next, a multi-layer perceptron (MLP) maps optical signals acquired from PDs to this learned latent space. Each reconstruction uses the measurement vector of a single LED scan, without memory from previous shape estimates.

\subsection{Problem Definition} \label{sec:data-represent}
Let $\boldsymbol{v} \in \mathbb{R}^{p\ell}$ be the measurement vector obtained by sequentially scanning $\ell$ LEDs and sampling the responses of $p$ photodiodes (Section~\ref{sec:electrical-system}), and $S = \left\{(x_i,y_i,z_i)\right\}_{i=1}^{M}\in\mathbb{R}^{M\times 3}$ be a point cloud of the sensor's surface geometry. The objective of shape reconstruction is defined as learning a mapping $f: \boldsymbol{v}\mapsto S_{\mathrm{pr}}$ that minimizes the discrepancy between the predicted surface $S_{\mathrm{pr}}$ and ground truth $S_{\mathrm{gt}}$ within the dataset $\mathcal{D}=\{(\boldsymbol{v}^{(i)},S_{\mathrm{gt}}^{(i)})\}_{i=1}^{N}$.

\subsection{Network Architecture}\label{sec:architecture}
\textbf{Stage 1: point-cloud autoencoder.}
The first stage learns a compact latent representation of the sensor's deformation space. As shown in Figure~\ref{fig:SI-model}A, given a ground-truth point cloud $S_{\mathrm{gt}}\in\mathbb{R}^{M_{\mathrm{gt}}\times 3}$, an encoder $E\left(\cdot\right)$ based on PointNet architecture~\cite{qi2017pointnet}  maps the input to a latent vector $\boldsymbol{z}\in\mathbb{R}^L$. A decoder $D\left(\cdot\right)$ composed of stacked transposed convolution layers then reconstructs a point cloud $S_{\mathrm{pr}}=D(\boldsymbol{z})\in\mathbb{R}^{M_{\mathrm{pr}}\times 3}$. PointNet performs a pointwise operation on an unordered point set, effectively extracting point cloud features. The autoencoder pre-training stage provides a shape prior that constrains reconstructions to lie on a learned manifold of feasible deformations.

\textbf{Stage 2: PD-to-latent regression.}
Once the latent representation of surface geometry is established, we freeze the autoencoder weights to preserve the learned shape prior. We then train an MLP $h(\cdot)$ to map the  measurement vector $\boldsymbol{v}$ to the latent space, such that $h(\boldsymbol{v})=\boldsymbol{z}$ (Figure~\ref{fig:SI-model}B). During inference, the predicted latent vector is passed through the pre-trained decoder to obtain the reconstructed shape, $S_{\mathrm{pr}}=D\left(\boldsymbol{z}\right)=D\left(h\left(\boldsymbol{v}\right)\right)$. This decomposition reduces the dimensionality of the regression target, transforming a high-dimensional coordinate prediction problem into a more tractable latent-space inference problem. Section~\ref{sec:SI-model} lists the layers of both networks.

\subsection{Training and Evaluation Metrics} \label{sec:train-eval-metric}
\textbf{Autoencoder training.}
The autoencoder is trained to reproduce its input point cloud using the squared Chamfer loss $\mathcal{L}_\mathrm{CD}$ (Equation~(\ref{eq:chamfer-loss})), a commonly adopted metric that quantifies the bidirectional similarity between the predicted and ground-truth point clouds~\cite{fan2017point, achlioptas2018learning}.
\begin{equation}
    \mathcal{L}_{\mathrm{CD}} = \sum_{\boldsymbol{x}\in S_{\mathrm{pr}}}\min_{\boldsymbol{y}\in S_{\mathrm{gt}}}\|\boldsymbol{x}-\boldsymbol{y}\|^2+ \sum_{\boldsymbol{y}\in S_{\mathrm{gt}}}\min_{\boldsymbol{x}\in S_{\mathrm{pr}}}\|\boldsymbol{x}-\boldsymbol{y}\|^2.
    \label{eq:chamfer-loss}
\end{equation}

\textbf{PD-to-latent regression training.}
For each training sample, the target latent vector is first computed as $\boldsymbol{z}=E\left(S_{\mathrm{gt}}\right)$. The MLP is then trained to minimize the mean squared error (MSE) between the predicted and target latent vectors, as given in Equation~(\ref{eq:MSE-loss}).
\begin{equation}
    \mathcal{L}_{\mathrm{mse}}\left(h(\boldsymbol{v}),\boldsymbol{z}\right)=\|h(\boldsymbol{v})-\boldsymbol{z}\|^2
    \label{eq:MSE-loss}
\end{equation}
Section~\ref{sec:SI-training} lists the optimizers, schedules, and other training settings.

\textbf{Hyperparameter selection.}
Both stages are trained on the same training frames, and all selection decisions use only the validation frames, which are disjoint from the held-out test intervals. For each stage, we retain the checkpoint with the lowest validation loss. The output resolution $M_{\mathrm{pr}}$ and the latent dimension $L$ are chosen from a sweep over twelve configurations evaluated on the validation set (Table~\ref{tab:SI-val-sweep}).

\textbf{Evaluation metrics.}
Reconstruction performance is evaluated on the test dataset using the Chamfer distance $d_{\mathrm{CD}}$, defined as the sum of the two mean Euclidean nearest-neighbor distances (Equation~(\ref{eq:chamfer-eval})), expressed in millimeters.

\begin{equation}
    d_{\mathrm{CD}}(S_{\mathrm{gt}}, S_{\mathrm{pr}}) = \frac{1}{M_{\mathrm{pr}}}\sum_{\boldsymbol{x}\in S_{\mathrm{pr}}}\min_{\boldsymbol{y}\in S_{\mathrm{gt}}}\|\boldsymbol{x}-\boldsymbol{y}\|_2 + \frac{1}{M_{\mathrm{gt}}}\sum_{\boldsymbol{y}\in S_{\mathrm{gt}}}\min_{\boldsymbol{x}\in S_{\mathrm{pr}}}\|\boldsymbol{x}-\boldsymbol{y}\|_2.
    \label{eq:chamfer-eval}
\end{equation}

In addition to this global metric, we assess local reconstruction accuracy using a spatial error map computed as the nearest-neighbor (NN) distance from each predicted point to the ground-truth point set, defined for $\boldsymbol{x}\in S_{\mathrm{pr}}$ in Equation~(\ref{eq:nnd}).
\begin{equation}
    d_{\mathrm{NN}}(\boldsymbol{x}) = \min_{\boldsymbol{y}\in S_{\mathrm{gt}}}\|\boldsymbol{x}-\boldsymbol{y}\|_2
    \label{eq:nnd}
\end{equation}
This visualization reveals spatial regions where reconstruction errors are most pronounced.
\section{Experiments and Results}
\begin{figure}[!htbp]
    \centering
    \includegraphics{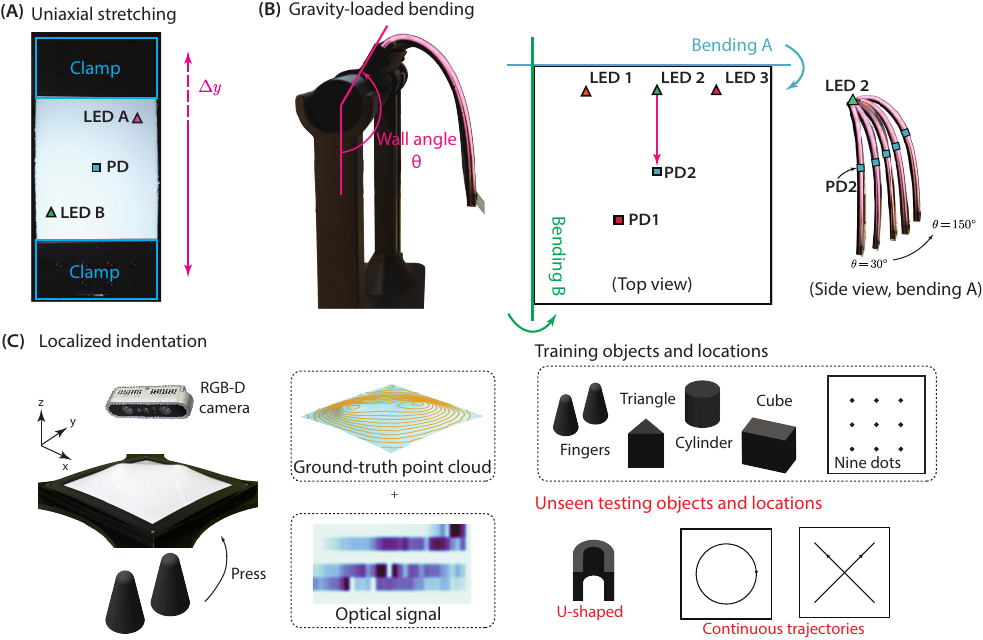}
    \caption{\textbf{Experimental setups.} \textbf{(A)} Uniaxial stretching of a 120~mm $\times$ 100~mm prototype with 7 LEDs and 1 PD (not all drawn). \textbf{(B)} Gravity-loaded bending of the 140~mm $\times$ 140~mm membrane with 3 LEDs and 2 PDs, clamped along the LED edge (bending A, along LED2--PD2) or the perpendicular edge (bending B). The wall angle $\theta$ sets the bending shape. \textbf{(C)} Localized indentation of the 140~mm $\times$ 140~mm membrane with 30 LEDs and 6 PDs clamped on all four sides, with an RGB-D camera providing the ground-truth point cloud. We pressed the training indenters at nine locations. The U-shaped indenter and the circular and cross-shaped finger trajectories were used for testing.}
    \label{fig:setup}
\end{figure}

We evaluate the membrane sensor through a series of experiments (Figure~\ref{fig:setup}). First, we characterize how global deformation modulates the optical signals and how stable these signals are over time. We then evaluate 3D shape reconstruction under localized indentation (Figure~\ref{fig:setup}C) and in-plane shape reconstruction under uniaxial stretching (Figure~\ref{fig:setup}A). Finally, we perform a feature ablation study using Shapley additive global importance (SAGE) values \cite{shapley1953value,covert2020understanding} to identify the redundancy among optoelectronic components and to provide design guidelines for future implementations.

\subsection{Optical Response to Global Deformation} \label{sec:response}
We first examine how global deformation modulates the light reaching the PDs, using two setups. In uniaxial stretching (Figure~\ref{fig:setup}A), the 120~mm $\times$ 100~mm membrane is clamped at both ends and stretched on a tensile testing machine (Instron) by a prescribed displacement $\Delta y$. In gravity-loaded bending (Figure~\ref{fig:setup}B), one edge of a 140~mm $\times$ 140~mm membrane is clamped, and the rest of the membrane bends under its own weight, with the clamp wall angle $\theta$ setting the bending shape. We follow the pair LED2--PD2 with the membrane clamped either along the LED edge (bending A), so that the LED--PD line runs along the bending direction, or perpendicular to the LED edge (bending B), so that the same line runs transverse to it.

\begin{figure}[!htbp]
    \centering
    \includegraphics{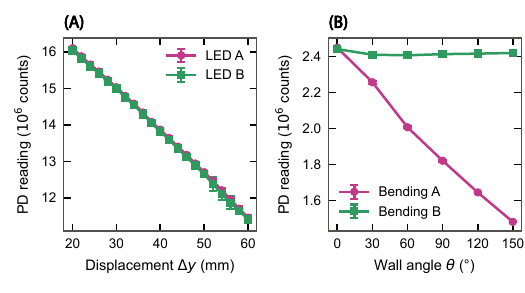}
    \caption{\textbf{Optical response to bending and stretching.} \textbf{(A)} PD reading against the displacement $\Delta y$ for two LEDs. \textbf{(B)} Reading of LED2--PD2 against the wall angle $\theta$ of the bending setup (Figure~\ref{fig:setup}B), with the LED--PD line along the bending (bending A) or transverse to it (bending B). Points show the mean, and error bars show the standard deviation of the one-second averages.}
    \label{fig:response}
\end{figure}

Stretching from $\Delta y=20$ to $60$~mm reduced the PD readings  for both LEDs monotonically (Figure~\ref{fig:response}A). Under bending angles from $0^\circ$ to $150^\circ$, the reading of LED2--PD2 decreased by 40\% in bending A, whereas it changed by less than 1\% in bending B (Figure~\ref{fig:response}B). The same LED--PD pair therefore responds differently depending on the orientation of its light path relative to the bending. This directional sensitivity indicates that the information content of the optical signals depends on the \textit{diversity of path orientations} through the waveguide, which motivates placing the LEDs along the edge and distributing the PDs over the central region (Section~\ref{sec:mechanical-system}).

We examined the effect of room light on the optical readings with the bending prototype (Figure~\ref{fig:setup}B). The same six wall angles from $0^\circ$ to $150^\circ$ were recorded under normal laboratory lighting and in a dark room. At each angle, the readings under the two lighting conditions differed by $-3.2$ to 2.4 times the typical fluctuation of the one-second averages, whereas bending over the full range changed the two most responsive channels by 34 and 85 times this fluctuation (Figure~\ref{fig:SI-lighting}, Table~\ref{tab:SI-lighting}). The effect of room light on these readings is therefore small compared with the deformation-induced signal changes.

\subsection{Localized Indentation Reconstruction}\label{sec:indentation}
We evaluate 3D shape reconstruction on the 140~mm $\times$ 140~mm membrane clamped on all four sides (Figure~\ref{fig:setup}C), where localized indentation produces large out-of-plane deformation. A depth camera (RealSense D435, Intel) facing the membrane records its surface at 30~Hz. We convert each depth frame into a ground-truth point cloud of 5{,}929 points and pair it with the most recent optical measurement (Section~\ref{sec:SI-electrical}).

\begin{figure}[!htbp]
    \centering
    \includegraphics[width=\linewidth]{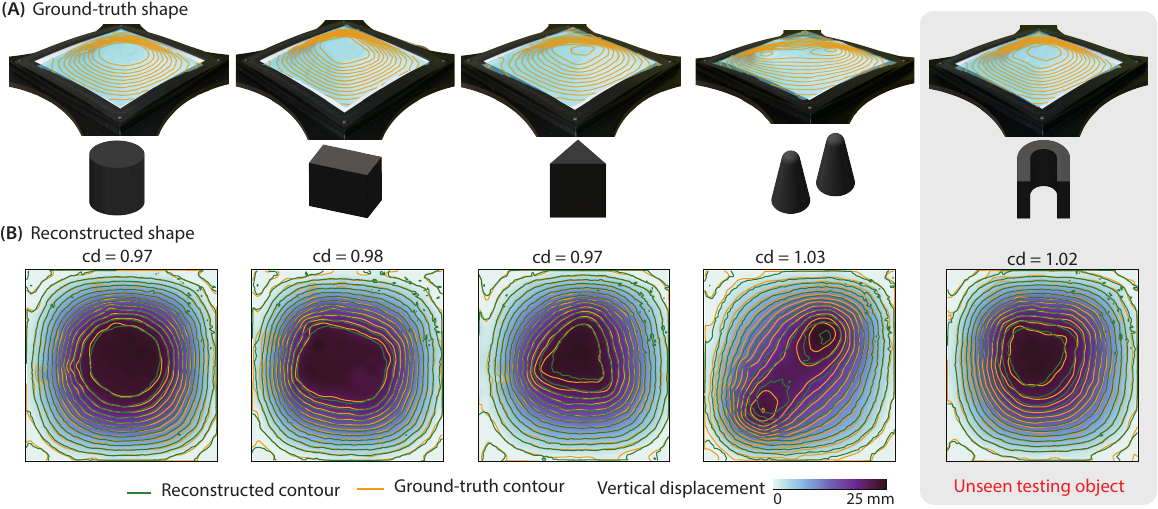}
    \caption{\textbf{Shape reconstruction for the five held-out indentation intervals.} \textbf{(A)} Ground-truth surface of the displayed frame, visualized as height contours on the sensor, above the indenter used. \textbf{(B)} Height contours of the reconstructed surface (green) overlaid on those of the ground truth (orange) and on the ground-truth vertical displacement (color). $d_{\mathrm{CD}}$ (cd, in mm) of each frame is given above it.}
    \label{fig:indentation}
\end{figure}

\subsubsection{Training Data and Evaluation Protocol} \label{sec:dataset}
To generate diverse surface geometries, we pressed fingers and rigid indenters with cylindrical, cuboid, and triangular tips into the membrane by hand at nine locations arranged in a $3\times3$ grid, with different orientations and depths (Figure~\ref{fig:setup}C). The recordings contain about 350{,}000 paired frames. We describe a frame's deformation magnitude by $\Delta z=z_{\max}-z_{\min}$ in its ground-truth point cloud. Across the training and validation frames, $\Delta z$ has a median of 15.1~mm, and 95\% and 99\% of frames are below 21.3~mm and 24.0~mm, respectively.

For testing, we recorded five separate pressing sessions of about 30~s each: one each with the cylinder, the cube, the triangle, and two fingers (also present in the training data), and one with a U-shaped indenter (not present). Each session begins and ends with the membrane undeformed and contains only complete presses. These five sessions contribute 4{,}512 test frames, and Section~\ref{sec:SI-dataset} details the recordings and the test sessions. To evaluate how precisely the reconstruction locates contact, we further recorded two continuous finger-stroke trajectories, a circle and a cross.

\begin{figure}[!htbp]
    \centering
    \includegraphics[width=\linewidth]{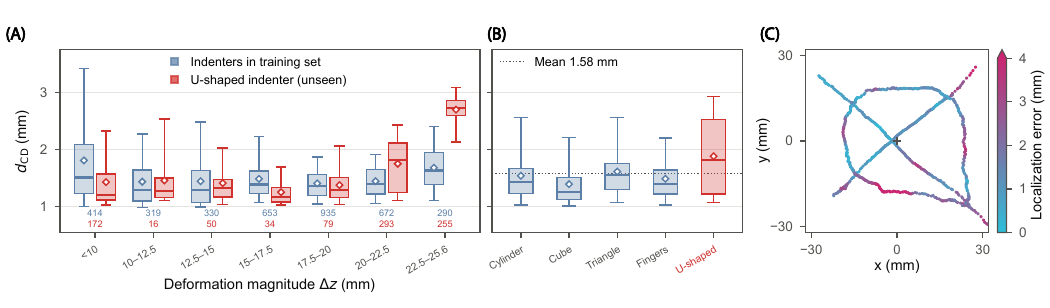}
    \caption{\textbf{Reconstruction error and localization error on the held-out test data.} \textbf{(A)} Chamfer distance $d_{\mathrm{CD}}$ against the deformation magnitude $\Delta z$ of the ground truth, for the four indenters represented in the training data (blue) and the U-shaped indenter absent from training (red). Numbers give the frames in each bin. \textbf{(B)} $d_{\mathrm{CD}}$ of each held-out interval. The dotted line marks the mean over all 4{,}512 frames. \textbf{(C)} Localization error along the cross and circle strokes. Each point is one frame, drawn at the measured contact position and colored by its localization error.}
    \label{fig:indentation-stats}
\end{figure}

\subsubsection{Reconstruction Performance}
Figure~\ref{fig:indentation} compares reconstructed and ground-truth surfaces for one deeply indented frame of each held-out interval. The reconstructed height contours follow the ground truth across the whole membrane and reproduce the outline of the contact region, including the indenter's footprint and the two separate finger contacts. Supplementary Video~1 shows the reconstruction running in real time while the membrane is pressed, including continuous finger trajectories along a circle and a cross that cover contact locations between the nine training points.

Over all 4{,}512 held-out frames, the mean $d_{\mathrm{CD}}$ is 1.580~mm, with a median of 1.427~mm and a standard deviation of 0.539~mm (Figure~\ref{fig:indentation-stats}). Table~\ref{tab:SI-test-intervals} and Figure~\ref{fig:SI-test-sequences} show the reconstruction results in detail. For the four indenters in the training data, the mean interval error ranges from 1.396~mm for the cube to 1.614~mm for the triangle (Figure~\ref{fig:indentation-stats}B). Their error depends little on the deformation magnitude up to $\Delta z=22.5$~mm, with a median between 1.29 and 1.51~mm in every bin (Figure~\ref{fig:indentation-stats}A). Above 22.5~mm, the median rises to 1.63~mm.

For the unseen U-shaped indenter, the mean error over its interval is 1.884~mm. This difference from the familiar indenters originates from the deepest frames. Below $\Delta z=20$~mm, the median error of the 351 U-shaped frames lies between 1.17 and 1.33~mm per bin, within the range of the familiar indenters. Between 20 and 22.5~mm it rises to 1.82~mm, and above 22.5~mm to 2.73~mm, compared with 1.63~mm for the familiar indenters at the same depth (Figure~\ref{fig:indentation-stats}A). Retraining the same model configuration with five different random seeds (Table~\ref{tab:SI-seeds}) gave mean errors of 1.54 to 1.64~mm over all test frames and of 1.58 to 1.91~mm for the U-shaped indenter. The membrane therefore reconstructs an indenter shape absent from training with an accuracy close to that of the training indenters for $\Delta z$ below 20~mm, whereas deep indentation by the unseen shape produces the largest errors.

\subsubsection{Contact Localization}
Figure~\ref{fig:cross} and Figure~\ref{fig:circle} show the reconstruction performance along the cross and circle strokes.
To evaluate how precisely the reconstruction model generalizes to unseen contact locations, we compare the contact position on the reconstructed surface with that on the ground-truth surface. Across both strokes, the mean localization error is 1.779~mm, about 1.3\% of the 140~mm membrane width (Figure~\ref{fig:indentation-stats}C). The mean $d_{\mathrm{CD}}$ over the stroke frames is 1.486~mm, close to that of the press intervals. Figure~\ref{fig:SI-stroke-timeseries} and Table~\ref{tab:SI-strokes} show the detailed statistics of contact localization error.

\subsection{Signal Stability under Repeated Deformation} \label{sec:stability}
We assessed the stability of the optical signals on the indentation membrane in Section~\ref{sec:indentation}, using 30 LEDs and 6 PDs, under two kinds of repeated deformation. First, over three consecutive sessions, we pressed the membrane repeatedly by hand while the depth camera recorded its surface. Second, we stretched the same membrane by different magnitudes repeatedly on a tensile test machine. Section~\ref{sec:SI-stability} details the experimental setup.

\begin{figure}[!htbp]
    \centering
    \includegraphics[width=\linewidth]{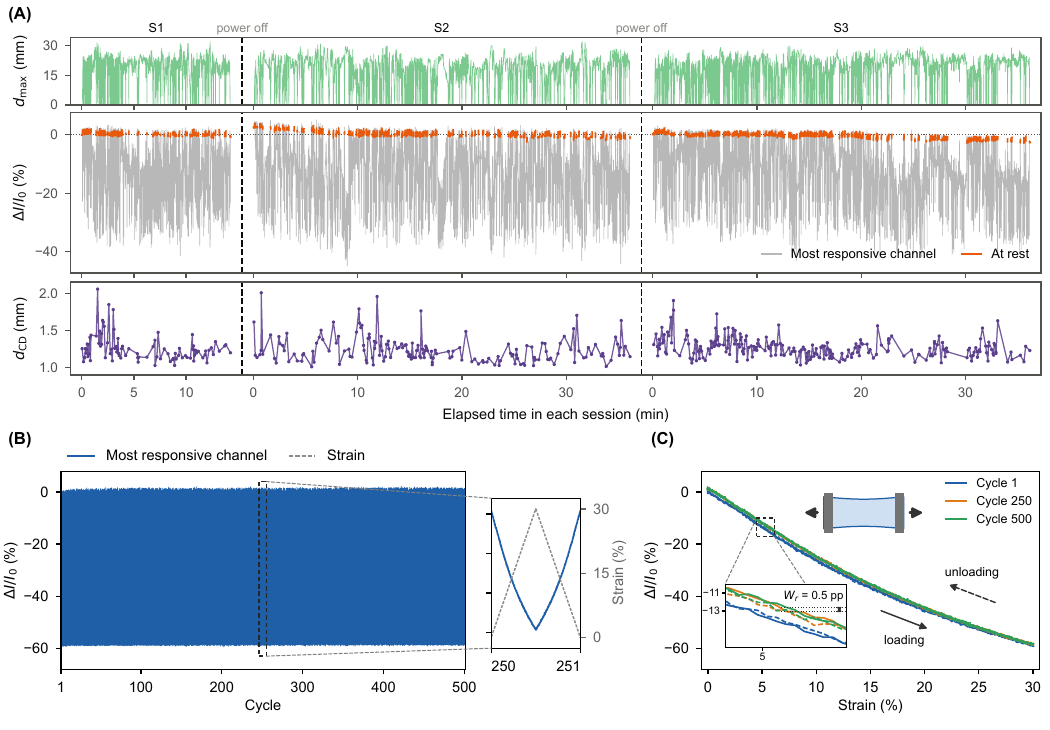}
    \caption{\textbf{Signal drift and hysteresis of the indentation membrane.} \textbf{(A)} Peak deformation $d_{\max}$ (top), sensor response of the most responsive channel (middle), and reconstruction error $d_{\mathrm{CD}}$ at rest (bottom) over three sessions with power cycling. Each dot is one rest period. \textbf{(B)} $\Delta I/I_0$ of the most responsive channel over 500 cycles of uniaxial stretching to 30\% strain, and  an enlarged view of one specific stretch cycle plotted with strain. \textbf{(C)} $\Delta I/I_0$ against strain in cycles 1, 250, and 500 on loading (solid) and unloading (dashed) branches. The  inset marks the loop width $W_r$ for cycle 500.}
    \label{fig:characterization}
\end{figure}

\subsubsection{Signal Drift}
We observe a slight drift of signals during approximately the first 10~min after power-on. Figure~\ref{fig:characterization}A shows the most responsive channel's reading relative to its reading at rest ($\Delta I/I_0$) over 471 manual presses across three sessions. After each power-on, the undeformed readings started above the reference and settled within approximately 10~min, with a maximum excursion of 3.5\%. The optical signals show no accumulated drift across the three sessions. The behavior is consistent with a decrease in LED output as the LEDs warm up.  Because the LED temperature was not measured, we leave confirming the cause of this start-up drift for future work. During the first 10~min after power-on, the warm-up drift increases the reconstruction error by about 5\% relative to the remainder of the session (Figure~\ref{fig:characterization}A, Table~\ref{tab:SI-sessions}).

\subsubsection{Repeatability}

We further evaluate the repeatability and validate the robustness of the liquid-metal circuits using a cyclic uniaxial stretching experiment. We clamped the membrane at two opposite edges and stretched it uniaxially to 30\% strain for 500 cycles while recording all 180 channels. We set the cycle strain slightly above the typical local strain of 28.2\% in the indentation, as estimated using finite element analysis (Figure~\ref{fig:SI-fea-strain}). Results show that the deformation response is consistent across cycles, with negligible long-term drift (Figure~\ref{fig:characterization}B). The liquid-metal circuit is also robust, with all 180 channels remaining functional after prolonged cyclic loading. The loading and unloading curves of cycles 1, 250, and 500 nearly overlap, with loop width $W_r$ of about 0.5 percentage points for the most responsive channel (Figure~\ref{fig:characterization}C). Figures~\ref{fig:SI-cyclic-channels} and~\ref{fig:SI-cyclic-channels2} show all channels' response during this cyclic test.

\subsection{Uniaxial Stretching Reconstruction} \label{sec:stretch-recon}
We further evaluate the shape reconstruction under extreme in-plane strain using the 120~mm $\times$ 100~mm prototype on the tensile testing machine (Figure~\ref{fig:setup}A). In the uniaxial stretching experiment, we captured ground truth with an RGB camera (Sony Alpha~7). The camera was mounted on a tripod with its optical axis perpendicular to the membrane plane, so the imaging is approximately orthographic and a single in-plane scale relates image and metric coordinates. For each frame, the membrane region is segmented in hue--saturation--value (HSV) space (bright, low-saturation pixels), refined by morphological closing and opening, and reduced to its largest connected component. We then map the segmented pixels from image coordinates to metric $(x,y)$ coordinates using a similarity transform derived from a two-point calibration with in-plane fixture points separated by a known distance of 100~mm. We set the out-of-plane coordinate to $z=0$, so the ground-truth and predicted point clouds share the same 3D metric coordinate system, and the reconstruction reduces to in-plane geometry.

\begin{figure}[!htbp]
    \centering
    \includegraphics{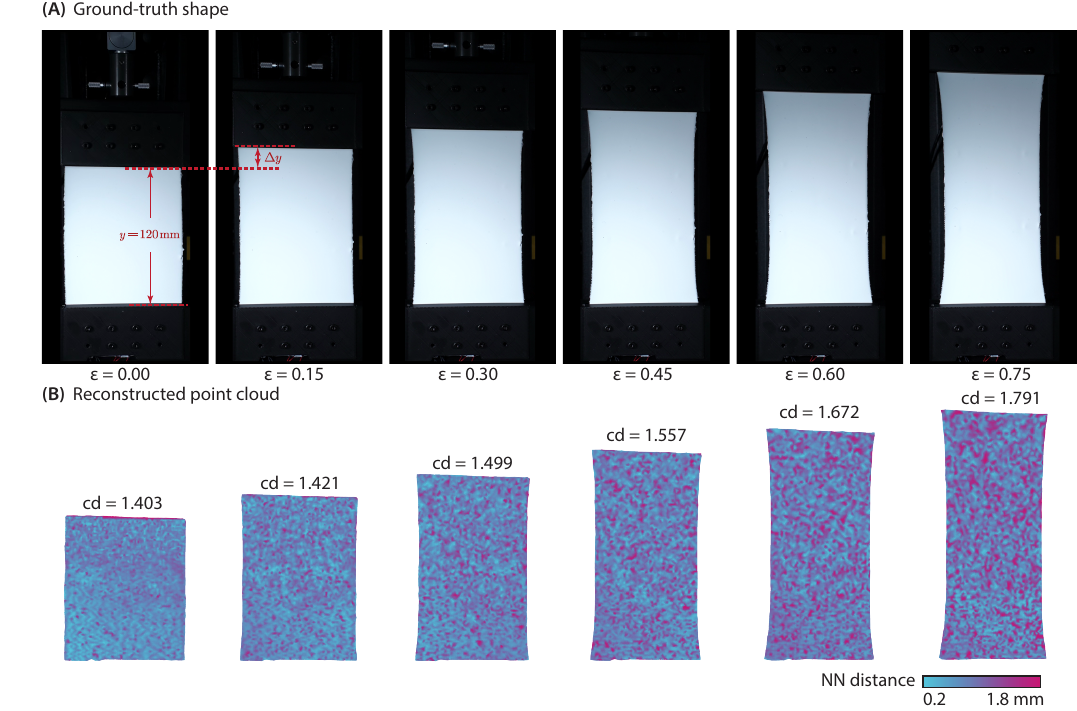}
    \caption{\textbf{Reconstruction performance under uniaxial stretching.} \textbf{(A)} Ground-truth geometries at selected stretching states, parameterized by the applied displacement $\Delta y$ (initial gauge length $y=120$~mm) and the corresponding axial strain $\varepsilon=\Delta y/y$ ($\varepsilon=0.75$ corresponds to 75\% strain). \textbf{(B)} Reconstructed point clouds for the same states, visualized with a nearest-neighbor (NN) distance colormap. The Chamfer distance $d_{\mathrm{CD}}$ (cd, in mm) of each case is annotated. The ground truth and the reconstruction lie in the membrane plane ($z=0$).}
    \label{fig:stretch}
\end{figure}

To build the training dataset, we sweep $\Delta y$ from 0 to 90~mm (corresponding to 75\% strain) in 2~mm increments. At each stretch step, we hold the displacement and record 5{,}000 frames, paired with the corresponding ground-truth geometry for that step captured by the camera. Using the collected data, we train  a separate shape reconstruction model described in Section~\ref{sec:neural-network}. We then evaluate the learned mapping on a separate stretching trial that follows the same loading protocol but is not used for training. Supplementary Video~2 summarizes the evaluation process, and Figure~\ref{fig:stretch} shows qualitative results. For each test frame, we reconstruct the in-plane outline as a point cloud and visualize the spatial error using a nearest-neighbor (NN) distance colormap relative to the ground truth (Figure~\ref{fig:stretch}B). The reconstructions capture the global elongation and preserve the overall shape profile throughout the entire loading range.

Over all evaluated frames between 0 and 75\% strain, the mean $d_{\mathrm{CD}}$ is 1.557~mm. We observe a slight increase in $d_{\mathrm{CD}}$ as strain increases (Figure~\ref{fig:stretch}B), from 1.403~mm without stretching to 1.791~mm at 75\% strain, and the nearest-neighbor distance increases across the membrane. These results indicate that the optical signal can still encode strain-induced shape information up to 75\% strain, close to the mechanical limit of our liquid-metal-based circuits (Figure~\ref{fig:SI-stretch-failure}).

\subsection{Feature Importance and Sensor Optimization} \label{sec:optimization}
The number of LEDs and PDs determines how many independent light-intensity measurements the membrane provides, but more components also increase fabrication complexity and readout time (Section~\ref{sec:hardware-design}). To guide future layouts, we estimated each LED and PD's contribution to reconstruction using group Shapley additive global importance (SAGE) values~\cite{covert2020understanding, covert2021explaining}, and retrained the PD-to-latent network with progressively larger subsets of LEDs or PDs (Figure~\ref{fig:optimization}). The importance of the components depends strongly on their position. The PD near the center of the membrane and many LEDs contribute little, which we attribute to a low baseline intensity at the PD caused by a large LED--PD distance or voltage drops along the LED power traces; small deformation-induced changes are then hard to resolve against noise. When LEDs are added in order of importance, the reconstruction error first decreases quickly and then levels off, so a subset of the LEDs already provides most of the information. The LED count can therefore likely be reduced with little loss in accuracy, whereas the PDs should be placed so that their baseline intensity stays within a favorable range.
\section{Conclusion}
We present a stretchable proprioceptive membrane that reconstructs its own surface from embedded optical measurements. Edge-mounted LEDs and centrally distributed PDs are embedded in a multilayer elastomeric waveguide and connected through liquid-metal interconnects with flexible PCB breakouts, so that the circuit deforms together with the elastomer. Time-multiplexed readout provides LED--PD measurements, and a two-stage neural network maps each set of intensity measurements to a dense point cloud of the membrane surface.

Under localized indentation, the membrane reconstructs surfaces with millimeter-level accuracy and localizes finger contacts to about 1.3\% of its width. Under uniaxial stretching up to the mechanical limit of the liquid-metal circuit, the optical signals still encode sufficient shape information for in-plane outline reconstruction. Including network inference, the reconstruction runs in real time at about 90~Hz.  Characterization of the optical signals shows a small start-up drift that settles after a short warm-up and hysteresis within the measurement noise. External light  has a small effect on optical readings, yet its effect on reconstruction performance was not evaluated. Together, these results indicate that a stretchable two-dimensional waveguide membrane can serve as a proprioceptive shape sensor for localized indentation and in-plane stretching.

While the presented waveguide membrane sensor achieves accurate reconstructions, several design choices can be further optimized. The current LED--PD layout is primarily guided by empirical design intuition. Section~\ref{sec:optimization} partially addresses this by quantifying feature importance using SAGE, and we leave principled layout optimization for future work. Light attenuation inside the waveguide bounds the usable LED--PD separation distance, which limits the size of a single sheet. Larger coverage  could be achieved by increasing source brightness or tiling multiple units. The liquid-metal circuit currently limits the maximum tolerable strain. Adopting serpentine interconnects could extend the operating range. Whether the membrane distinguishes the two opposite directions of out-of-plane deformation also remains open.

Beyond standalone sensor characterization, integrating this membrane into robotic systems opens several avenues for future work. Deploying the sensor onto soft robots or robotic grippers could enable closed-loop proprioceptive control and contact-aware manipulation. In addition, extending the sensing capability from pure geometric reconstruction to distributed force estimation would broaden its utility as a robotic tactile sensor. Specifically, spatial force distributions could be inferred from the reconstructed 3D surface geometry via mechanical modeling. Alternatively, the two-stage framework could be trained directly with force-field supervision, where the autoencoder learns a latent representation of contact force distributions, and the MLP maps optical intensity measurements to this representation.

\medskip
\noindent\textbf{Acknowledgments} \par
G.X., J.W., D.T., and X.H. acknowledge support from the Office of Naval Research (ONR) under Grant No.~N000142412137 and from the National Science Foundation (NSF) under Grant No.~IIS-2332554. The authors also thank Y.M. and M.V. for suggestions on fabrication. The authors are grateful to other members of the research group for their help with this work.

\clearpage
\pdfbookmark[1]{Supporting Information}{msSI}
\section*{Supporting Information}
\setcounter{subsection}{0}
\renewcommand{\theHsubsection}{SI.\arabic{subsection}}
\renewcommand{\thesubsection}{S\arabic{subsection}}
\setcounter{figure}{0}
\renewcommand{\thefigure}{S\arabic{figure}}
\renewcommand{\theHfigure}{SI.\arabic{figure}}
\setcounter{table}{0}
\renewcommand{\thetable}{S\arabic{table}}
\renewcommand{\theHtable}{SI.\arabic{table}}
\setcounter{equation}{0}
\renewcommand{\theequation}{S\arabic{equation}}
\renewcommand{\theHequation}{SI.\arabic{equation}}

Supporting Information includes Supplementary Video 1 (real-time indentation reconstruction), Supplementary Video 2 (uniaxial stretching reconstruction), and the supplementary text, figures, and tables appended below.
\subsection{Material Selection and Fabrication Details}\label{sec:SI-fabrication-detail}
\subsubsection{Materials}
The transparent core is made of Ecoflex 00-45 Near Clear (Smooth-On, Inc.), a silicone rubber that combines high optical transmittance with high compliance under large deformation. The reflective layers are a 3:1 (mass ratio) mixture of Ecoflex 00-45 and Print-On White Silicone Ink (Raw Material Suppliers). The cladding layers are a mixture of Ecoflex 00-45 and Print-On Black Silicone Ink (Raw Material Suppliers) (Figure~\ref{fig:hardware}A). We selected these formulations through iterative prototyping to achieve high reflectance in the reflective layers and strong attenuation in the cladding. The circuit substrate is 3M VHB 4905 tape, and the interconnects are oxidized gallium--indium (OGaIn)~\cite{woodman2024stretchable}.

\subsubsection{Fabrication Steps}
The membrane is fabricated layer by layer in a 3D-printed square mold with a width of 140~mm and a depth of 6~mm (Figure~\ref{fig:fabrication}).
\begin{enumerate}
    \item Spin-coat the bottom black cladding layer at 200~rpm and partially cure it at 60~$^\circ$C for 15~min (Figure~\ref{fig:fabrication}A).
    \item Spin-coat the bottom white reflective layer onto the black layer and fully cure it at 25~$^\circ$C for 240~min (Figure~\ref{fig:fabrication}B).
    \item Place a VHB 4905 sheet cut to 135~mm $\times$ 135~mm on the cured white layer. Seal the 2.5~mm margin on each side with Ecoflex 00-45 (Figure~\ref{fig:fabrication}C). A full-size VHB sheet tends to delaminate from the silicone when the liner is peeled, and the perimeter seal prevents this.
    \item Use a CO$_2$ laser cutter to cut the trace pattern into the release liner on the exposed face of the VHB sheet and peel off the cut regions of the liner to expose the adhesive (Figure~\ref{fig:fabrication}D).
    \item Paint OGaIn over the liner to fill the exposed pattern (Figure~\ref{fig:fabrication}E). Removing the remaining liner leaves the OGaIn traces on the adhesive (Figure~\ref{fig:fabrication}F).
    \item Check electrical continuity of every trace with a multimeter. Then press the flexible flat cable and the flexible PCB breakouts carrying the LEDs and PDs onto the traces at their designed locations, with the LED breakouts folded by 90$^\circ$ (Figure~\ref{fig:fabrication}G). The VHB pressure-sensitive adhesive holds each part in place, and no soldering is used.
    \item Pour the transparent core into the mold over the circuit and partially cure it (Figure~\ref{fig:fabrication}H).
    \item Cast the top white reflective layer and the top black cladding layer with the same formulations as the bottom layers (Figure~\ref{fig:fabrication}I--J).
    \item Demold the membrane after curing for two days.
\end{enumerate}
For the localized indentation and uniaxial stretching experiments, we additionally coated the outer surface of the membrane with a thin matte white layer to improve depth-camera readings and image segmentation.

\subsubsection{Optical Operating Range}
\begin{figure}[!htbp]
    \centering
    \includegraphics[width=0.5\linewidth]{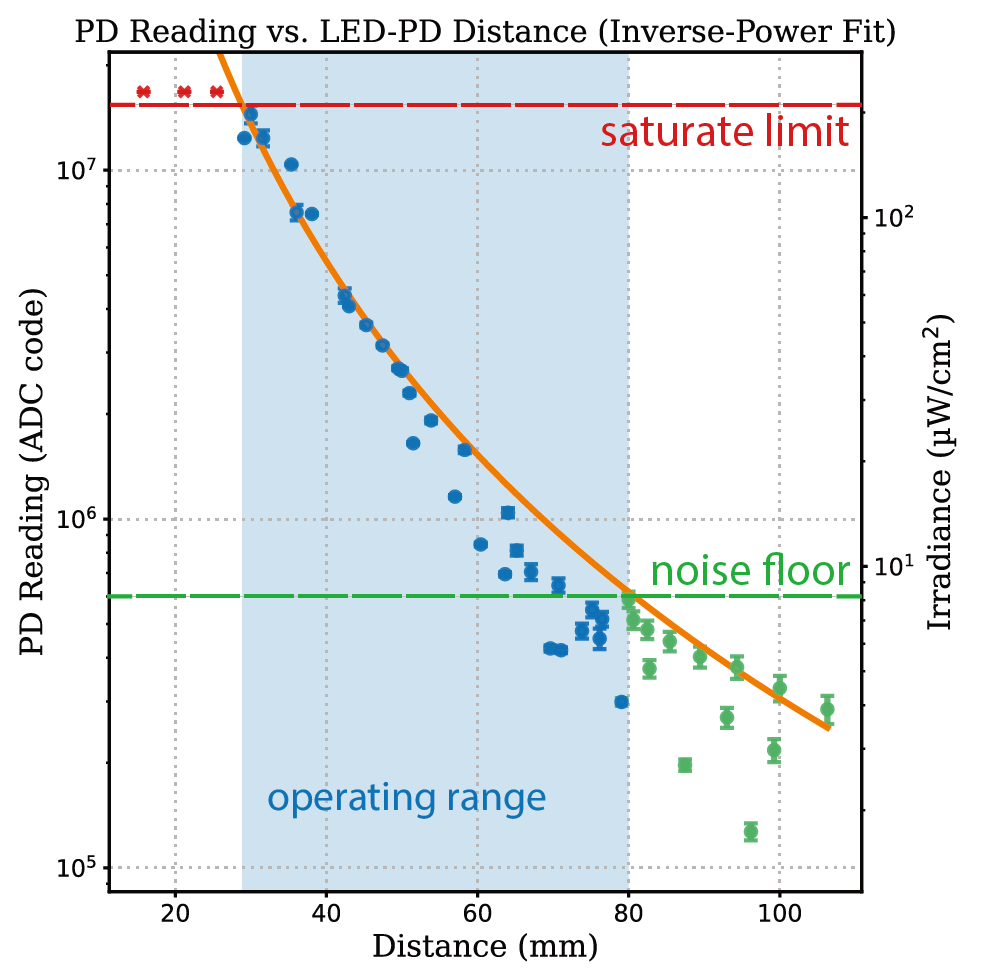}
    \caption{PD reading as a function of the LED--PD distance in the waveguide. Points show the mean reading of individual LED--PD pairs on a logarithmic scale, with error bars indicating the variation of repeated readings, and the orange curve is an inverse-power fit. The right axis gives the corresponding irradiance. Red crosses mark pairs that saturate the ADC, and green points mark pairs below the noise floor. The shaded band marks the operating range between the saturation limit and the noise floor, from about 29~mm to about 80~mm.}
    \label{fig:SI-pdreadings}
\end{figure}

Figure~\ref{fig:SI-pdreadings} shows how the PD reading depends on the LED--PD distance. The reading decreases steeply with distance and follows an inverse-power trend. Pairs closer than about 29~mm saturate the ADC, and pairs farther than about 80~mm fall below the ADC noise floor. This operating range guided the LED--PD layout described in Section~\ref{sec:mechanical-system}.

\subsection{Electronics and Data Acquisition}\label{sec:SI-electrical}
\begin{table}[!htbp]
    \centering
    \caption{Electronic components of the 140~mm $\times$ 140~mm prototype.}
    \label{tab:SI-components}
    \begin{adjustbox}{max width=\linewidth}
\begin{tabular}{>{\raggedright\arraybackslash}p{0.2\linewidth}>{\raggedright\arraybackslash}p{0.3\linewidth}>{\raggedright\arraybackslash}p{0.42\linewidth}}
        \toprule
        Function & Component & Configuration \\
        \midrule
        Light source & SK9822-EC20 (Opsco Optoelectronics) & Brightness 31/31, white (R+G+B), SPI clock 30~MHz \\
        Detector & ADPD2211 (Analog Devices) & 100 k$\Omega$ load resistor \\
        Digitization & AD7175-8 (Analog Devices) & 2.5~V reference, unipolar, SPI clock 15~MHz \\
        Controller & NUCLEO-L432KC (STMicroelectronics) & \\
        Host link & UART & 2~Mbaud \\
        Connector & Flexible flat cable & 20 pins at 2~mm pitch \\
        \bottomrule
    \end{tabular}
\end{adjustbox}
\end{table}
Table~\ref{tab:SI-components} lists the electronic components. The 24-bit ADC resolves small intensity variations at PDs far from the active LED, where the signal is much lower than at nearby channels. Each PD requires its own analog signal line, whereas the LED chain shares its bus lines, so the flexible flat cable's 20 usable conductors limit the number of PDs.

\subsubsection{Optical Signal Acquisition} \label{sec:SI-acquisition}
In each frame, the microcontroller records the PD readings once with all LEDs off, then switches on LEDs 1 to 30 one at a time, recording all PD readings in each step. A frame therefore contains a no-light reading of each PD and the $30\times6=180$ LED--PD readings (Figure~\ref{fig:hardware}B). The microcontroller sends each frame to the host as one line of comma-separated values with a millisecond timestamp through a DMA-driven UART, and it starts the next scan while the transfer is still in progress. Across eight independent recordings, the median interval between consecutive frames received by the host was 11.0~ms. We refer to the corresponding rate of about 90~Hz as the acquisition rate.

\begin{figure}
    \centering
    \includegraphics[width=0.5\linewidth]{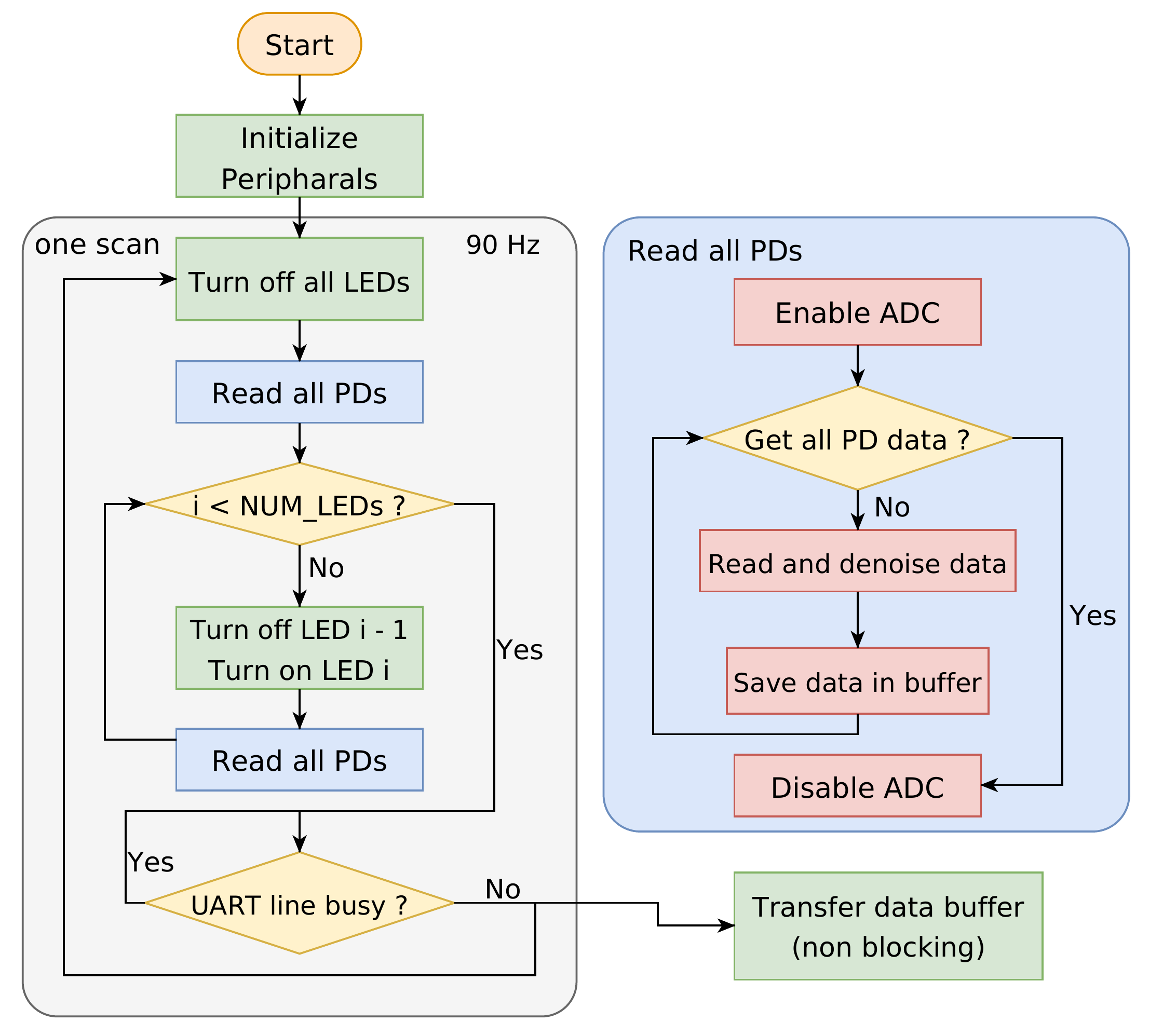}
    \caption{Data acquisition flowchart. Each scan records photodiode (PD) signals with all LEDs off, followed by sequential LED activation and PD readout. Buffered data are transmitted via non-blocking UART communication. The complete scan operates at 90~Hz.}
    \label{fig:flowchart}
\end{figure}

\subsubsection{Ground-Truth Shape Acquisition} \label{sec:SI-gt-acquisition}
In the localized indentation experiment, a single loop on the host waits for each depth frame from the RealSense D435 (Intel) depth camera at 30~Hz. It then reads all bytes waiting at the serial port, keeps the most recent complete optical frame, and saves it together with the depth frame as one paired sample. Optical frames that arrive between two depth frames are not stored. In the uniaxial stretching experiment, the tensile tester holds each displacement step while the sensor records 5,000 optical frames, and one camera image is taken per step. Every optical frame of a step is paired with the outline segmented from that image.

\subsubsection{Real-Time Reconstruction} \label{sec:SI-realtime}
The online reconstruction runs in three threads on the host. A receiving thread reads the serial port in blocks, splits the stream into complete frames, and keeps only the latest frame in a single-slot buffer. An inference thread takes this frame, discards any older one, extracts the 180 LED--PD readings, and evaluates $D(h(\boldsymbol{v}))$ on the laptop GPU (NVIDIA T1200 Laptop GPU). It places the predicted point cloud in a second single-slot buffer. A visualization thread renders the latest predicted point cloud whenever the display refreshes. Because each buffer holds only the latest item, a slower stage skips frames and does not accumulate delay. Acquisition, serial transfer, and inference are therefore pipelined. The forward pass takes 5 to 7~ms, shorter than the 11~ms frame interval, so the inference thread keeps pace with acquisition and reconstruction updates at about 90~Hz. This rate covers acquisition, serial transfer, preprocessing, and inference. Visualization runs asynchronously on the most recent prediction and is not included. Supplementary Video~1 is a screen recording at 30 frames per second, so it shows the reconstruction only qualitatively and does not by itself resolve the 90~Hz update rate.

\subsection{Implementation Details of the Neural Network} \label{sec:SI-nn}
\subsubsection{Network Architecture} \label{sec:SI-model}
\begin{figure}[!htbp]
    \centering
    \includegraphics[width=\linewidth]{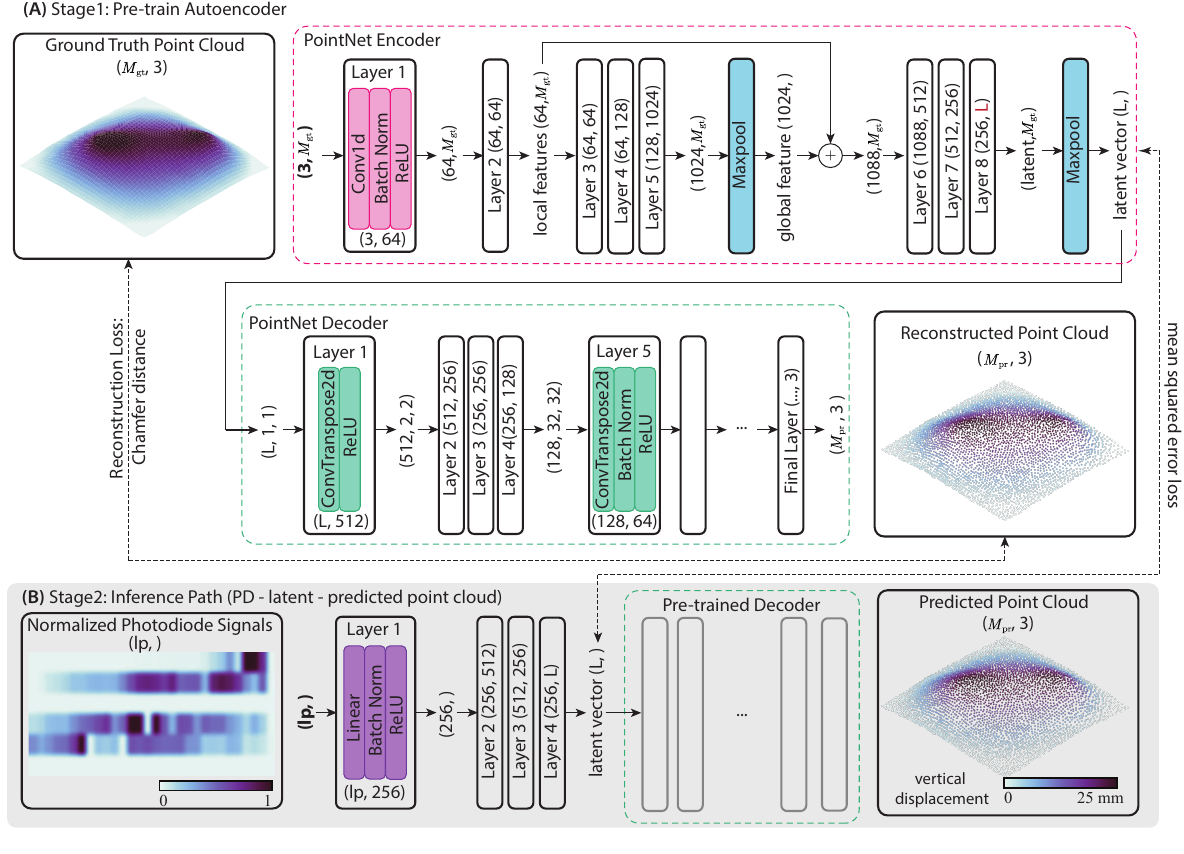}
    \caption{\textbf{Data-driven model for surface shape reconstruction.} \textbf{(A)} Stage 1: Autoencoder pre-training. A point-cloud autoencoder is trained using ground-truth surface point clouds. The encoder compresses the point cloud (shape: $(M_{\mathrm{gt}}, 3)$) into a latent vector (shape: $(L,)$), and the decoder reconstructs the point cloud (shape: $(M_{\mathrm{pr}}, 3)$) from the latent space. \textbf{(B)} Optical signal to shape inference. Normalized photodiode measurements (shape: $(p\ell,)$) are mapped to a latent vector using a multi-layer perceptron, and the pre-trained point-cloud decoder generates the predicted point cloud (shape: $(M_{\mathrm{pr}}, 3)$).}
    \label{fig:SI-model}
\end{figure}

Figure~\ref{fig:SI-model} shows the two networks. All hidden layers use batch normalization and ReLU activation unless stated otherwise, and all weights are initialized with the Xavier uniform scheme.

\paragraph{Encoder}
The encoder follows PointNet~\cite{qi2017pointnet} without the input and feature transform networks. Shared pointwise layers, implemented as one-dimensional convolutions with kernel size 1, map each point from 3 to 64 and 64 channels, forming the point's local feature. Three further shared layers map the local feature to 64, 128, and 1024 channels, and max pooling over all points gives a 1024-dimensional global feature. The global feature is concatenated with each local feature, and three shared layers map the resulting 1088 channels to 512, 256, and $L$ channels, with ReLU activation and no batch normalization in the last layer. A final max pooling over all points gives the latent vector $\boldsymbol{z}\in\mathbb{R}^{L}$.

\paragraph{Decoder}
The latent vector is reshaped into an $L\times1\times1$ tensor and upsampled by four two-dimensional transposed convolutions to $512\times2\times2$, $256\times4\times4$, $256\times10\times10$, and $128\times32\times32$. For $M_{\mathrm{pr}}=4096$, a further transposed convolution upsamples the feature map to $64\times64\times64$, and a $1\times1$ transposed convolution maps it to three channels. The resulting $3\times64\times64$ map is read out as 4096 points. The heads for $M_{\mathrm{pr}}=1024$ and $2048$ produce $32\times32$ and $32\times64$ grids in the same way.

\paragraph{PD-to-latent MLP}
We apply min-max normalization to each channel to account for signal-level differences across channels of $\boldsymbol{v}$. The MLP maps the normalized $p\ell$-dimensional measurement vector through fully connected layers with 256, 512, and 256 units to the $L$-dimensional latent vector. The output layer is linear.

\subsubsection{Network Training} \label{sec:SI-training}
\paragraph{Autoencoder}
We implement the networks in PyTorch~\cite{paszke2019pytorch} and train them on a single GPU with random seed 56. We train the autoencoder with the Adam optimizer~\cite{kingma2015adam} and a batch size of 16 for 120{,}000 steps, corresponding to about seven passes over the training frames. The loss $\mathcal{L}_{\mathrm{CD}}$ (Equation~(\ref{eq:chamfer-loss})) is computed per sample and averaged over the mini-batch. The learning rate increases linearly to $10^{-3}$ over the first 500 steps and then decays to $10^{-5}$ along a cosine schedule.
The batch-normalization momentum decays from 0.5 to 0.01 during training, and the running statistics are recomputed over 100 training batches before each validation. We evaluate the validation loss every 2{,}000 steps on 4{,}000 evenly spaced validation frames, and retain the checkpoint with the lowest validation loss.

\paragraph{PD-to-latent MLP}
The target latent vectors are computed once with the frozen encoder. We train the MLP with the Adam optimizer, a batch size of 64, and an initial learning rate of $10^{-3}$, which decays along a cosine schedule with $T_{\max}=100$ epochs. Training stops when the validation MSE has not improved for 20 epochs, and the checkpoint with the lowest validation MSE is retained. For the reported model, training stopped after 90 epochs, and we retained the checkpoint from epoch 70.

\subsubsection{Hyperparameter Selection}\label{sec:SI-sweep}
We selected the output resolution $M_{\mathrm{pr}}$ and the latent dimension $L$ on the validation set. For each of the twelve combinations of $M_{\mathrm{pr}}\in\{1024, 2048, 4096\}$ and $L\in\{128, 256, 512, 1024\}$, both stages were trained with the procedure of Section~\ref{sec:SI-training} and the same random seed, and the Chamfer distance of Equation~(\ref{eq:chamfer-eval}) was evaluated on all 69{,}096 validation frames. The ground-truth resolution is fixed at $M_{\mathrm{gt}}=5929$ for all configurations. Table~\ref{tab:SI-val-sweep} shows the results. The validation error decreases as $M_{\mathrm{pr}}$ increases for every $L$. At $M_{\mathrm{pr}}=4096$, the four latent dimensions give the same validation error at two decimal places, so $L$ has little effect. We therefore use $L=256$, the value adopted in our earlier experiments, together with $M_{\mathrm{pr}}=4096$ and $M_{\mathrm{gt}}=5929$. Each configuration was trained once.

\begin{table}[!htbp]
\centering
\caption{Validation Chamfer distance (mm) of each configuration, given as mean $\pm$ standard deviation. The mean is taken over all 69{,}096 validation frames.}
\label{tab:SI-val-sweep}
\begin{adjustbox}{max width=\linewidth}
\begin{tabular}{llcccc}
\toprule
 & & \multicolumn{4}{c}{Latent dimension $L$} \\
\cmidrule(lr){3-6}
$M_\mathrm{pr}$ & $M_\mathrm{gt}$ & 128 & 256 & 512 & 1024 \\
\midrule
1024 & 5929 & $2.32 \pm 0.11$ & $2.35 \pm 0.11$ & $2.32 \pm 0.11$ & $2.33 \pm 0.09$ \\
2048 & 5929 & $1.99 \pm 0.10$ & $1.98 \pm 0.11$ & $2.01 \pm 0.11$ & $1.98 \pm 0.11$ \\
4096 & 5929 & $1.43 \pm 0.14$ & $1.43 \pm 0.15$ & $1.43 \pm 0.14$ & $1.43 \pm 0.13$ \\
\bottomrule
\end{tabular}
\end{adjustbox}
\end{table}

\subsection{Optical Responses under Different Lighting Conditions}\label{sec:SI-lighting}

We clamped one edge of the 140~mm $\times$ 140~mm membrane in a holder, and the rest of the membrane bent under its own weight. Rotating the clamp set the wall angle $\theta$ and, in turn, the bending shape. We recorded two series, one under normal laboratory lighting and one in a dark room, at six wall angles from $0^{\circ}$ to $150^{\circ}$, each setting for about 60~s at about 1{,}000 frames per second. Three LEDs (L1--L3) and two PDs (P1--P2) were read out, giving six LED--PD channels, denoted L$i$P$j$ for LED $i$ and PD $j$. The per-frame no-light reading was subtracted from each channel. Each reading was averaged over 1~s after the first and last 5~s of each recording were discarded.

\begin{figure}[!htbp]
    \centering
    \includegraphics[width=0.6\linewidth]{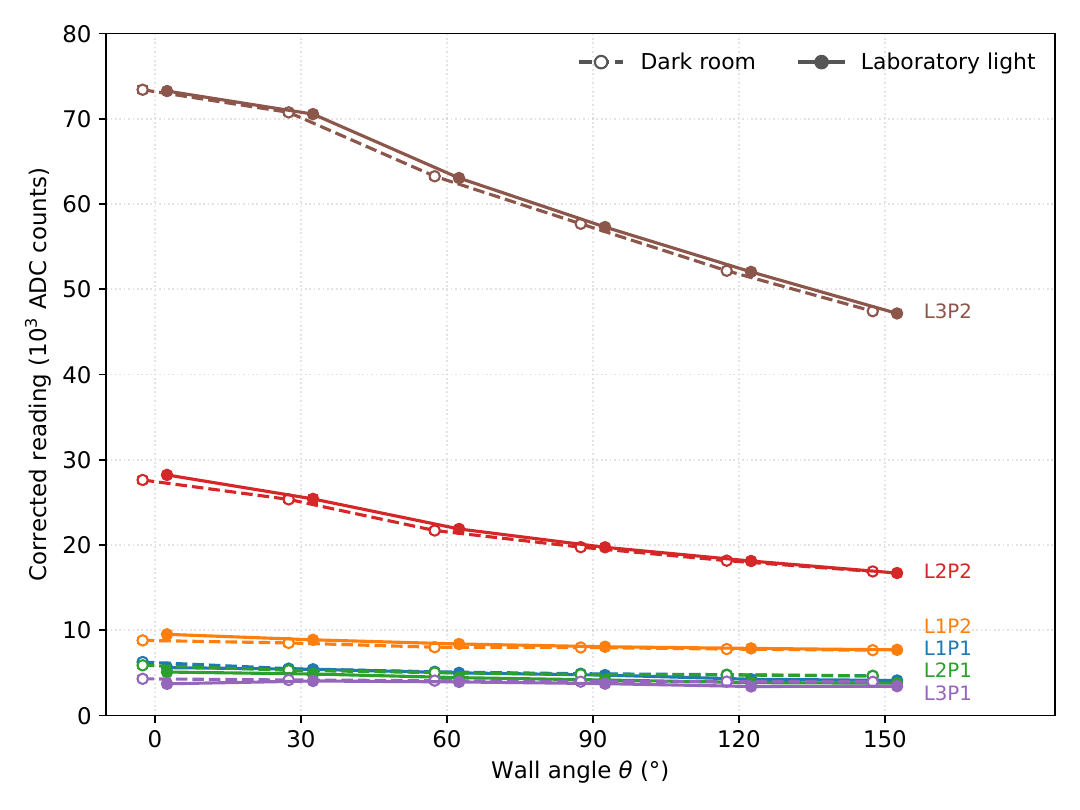}
    \caption{Readings of the bending arrangement at wall angles from $0^{\circ}$ to $150^{\circ}$, under laboratory lighting (solid lines, filled markers) and in a dark room (dashed lines, open markers). Markers show the mean over about 60~s of recording, and error bars show one standard deviation of the 1-s averages.}
    \label{fig:SI-lighting}
\end{figure}

Figure~\ref{fig:SI-lighting} shows the mean and standard deviation of the 1-s averages of each reading in each recording. The difference $\Delta I$ between the mean readings under the two lighting conditions at the same angle was expressed relative to the typical fluctuation $s_{\mathrm{typ}}=\sqrt{(s_{\mathrm{lab}}^2+s_{\mathrm{dark}}^2)/2}$ of the 1-s averages. $\Delta I/s_{\mathrm{typ}}$ ranged from $-3.2$ to 2.4 (Table~\ref{tab:SI-lighting}), while bending over the full range changed L2P2 and L3P2 by 34 and 85 times $s_{\mathrm{typ}}$. These results suggest that room light has a small effect on the readings compared with the response of L2P2 and L3P2 to bending.

Each 1-s average under laboratory lighting was assigned to the dark-room recording whose mean readings were nearest in Euclidean distance, among the six angles. All 299 averages were assigned to the correct angle. We did not evaluate the shape reconstruction model under different lighting conditions.

\begin{table}[!htbp]
\centering
\caption{Readings of the six LED--PD channels of the bending arrangement at $0^{\circ}$ to $150^{\circ}$ in ADC counts. Level: dark-room mean at $0^{\circ}$. Bending range: difference between the largest and smallest dark-room mean over the six angles. $s_{\mathrm{typ}}$: median over the six angles of the typical fluctuation of the 1-s averages. $\Delta I/s_{\mathrm{typ}}$: laboratory minus dark-room mean at the same angle relative to $s_{\mathrm{typ}}$, range over the six angles.}
\label{tab:SI-lighting}
\small
\begin{adjustbox}{max width=\linewidth}
\begin{tabular}{lrrrr}
\toprule
Channel & Level & Bending range & $s_{\mathrm{typ}}$ & $\Delta I/s_{\mathrm{typ}}$ \\
\midrule
L1P1 & 6{,}267 & 1{,}617 & 292 & $-2.36$ to $-0.24$ \\
L1P2 & 8{,}810 & 1{,}184 & 308 & 0.15 to 2.35 \\
L2P1 & 5{,}877 & 1{,}240 & 302 & $-3.20$ to $-1.44$ \\
L2P2 & 27{,}670 & 10{,}810 & 317 & $-0.71$ to 1.66 \\
L3P1 & 4{,}310 & 372 & 289 & $-2.13$ to $-0.42$ \\
L3P2 & 73{,}435 & 26{,}034 & 305 & $-1.29$ to $-0.43$ \\
\bottomrule
\end{tabular}
\end{adjustbox}
\end{table}

\subsection{Experiment Details of Indentation Reconstruction}\label{sec:SI-indentation}
\subsubsection{Dataset Collection}\label{sec:SI-dataset}
We recorded the indentation data using the 140~mm $\times$ 140~mm membrane in Figure~\ref{fig:setup}C. In manual-pressing sessions, we use cylinder, cube, triangle, and finger-shaped indenters to press the nine locations shown in Figure~\ref{fig:setup}C with different orientations and depths. This recording contains 345{,}571 frames. The deformation magnitude $\Delta z=z_{\max}-z_{\min}$ of the ground-truth point cloud has a median of 15.1~mm, an interquartile range of 11.3 to 17.7~mm, and 95th, 99th, and 99.9th percentiles of 21.3, 24.0, and 27.0~mm. Each ground-truth point cloud is taken from a fixed $230\times230$ pixel region of the depth image that covers the membrane. Sampling every third pixel ($s=3$) gives a $77\times77$ grid with $M_{\mathrm{gt}}=5929$ points.
In addition, we recorded press intervals for the cylinder, cube, triangle, fingers, and the U-shaped indenter of about 30~s each for testing. Each session begins and ends with the membrane undeformed and contains only complete presses. The training and validation data contain no presses of the U-shaped indenter, which appears only in its test interval. We further recorded two continuous finger strokes along a cross and a circle, which cover locations unseen during training.

\subsubsection{Press Segmentation and Data Split}
We sampled each depth frame on a $77\times77$ grid. We fitted a plane to the outer three rows and columns of the grid, which lie on the clamped boundary, and computed deformation relative to this plane and the session's resting profile. The resting profile is the median over the undeformed frames. We smoothed the deformation map with a Gaussian kernel of 2 grid cells, and $d_{\max}$ is the maximum over a $17\times17$ subsample of the map after trimming 4 cells from each edge. A frame is undeformed when $d_{\max}<1.5$~mm, and a rest period consists of at least 3 consecutive undeformed frames. A press comprises the frames between two rest periods and must reach $d_{\max}\geq 8$~mm. We use approximately the first 80\% (276{,}366 frames) for training and the remaining 20\% (69{,}096 frames) for validation, rounded to the nearest complete press interval.

\begin{table}[!htbp]
\centering
\caption{Held-out test intervals and reconstruction error of the reported model. $\Delta z$: median and maximum over the interval. $d_{\mathrm{CD}}$: mean, median, standard deviation, and 95th percentile over the frames of the interval.}
\label{tab:SI-test-intervals}
\small
\begin{adjustbox}{max width=\linewidth}
\begin{tabular}{lrrrrrrrr}
\toprule
Indenter & Frames & Duration & \multicolumn{2}{c}{$\Delta z$ (mm)} & \multicolumn{4}{c}{$d_{\mathrm{CD}}$ (mm)} \\
\cmidrule(lr){4-5}\cmidrule(lr){6-9}
 & & (s) & Median & Max & Mean & Median & SD & 95th pct. \\
\midrule
Cylinder & 817 & 28 & 19.1 & 22.0 & 1.543 & 1.430 & 0.457 & 2.571 \\
Cube & 1{,}066 & 36 & 16.8 & 24.3 & 1.396 & 1.255 & 0.449 & 2.218 \\
Triangle & 879 & 30 & 18.6 & 25.6 & 1.614 & 1.549 & 0.481 & 2.559 \\
Fingers & 851 & 28 & 17.5 & 25.3 & 1.490 & 1.396 & 0.480 & 2.200 \\
U-shaped (unseen) & 899 & 30 & 21.0 & 24.4 & 1.884 & 1.818 & 0.668 & 2.933 \\
\midrule
All & 4{,}512 & & 18.2 & 25.6 & 1.580 & 1.427 & 0.539 & \\
\bottomrule
\end{tabular}
\end{adjustbox}
\end{table}

\begin{figure}[!htbp]
    \centering
    \includegraphics[width=\linewidth]{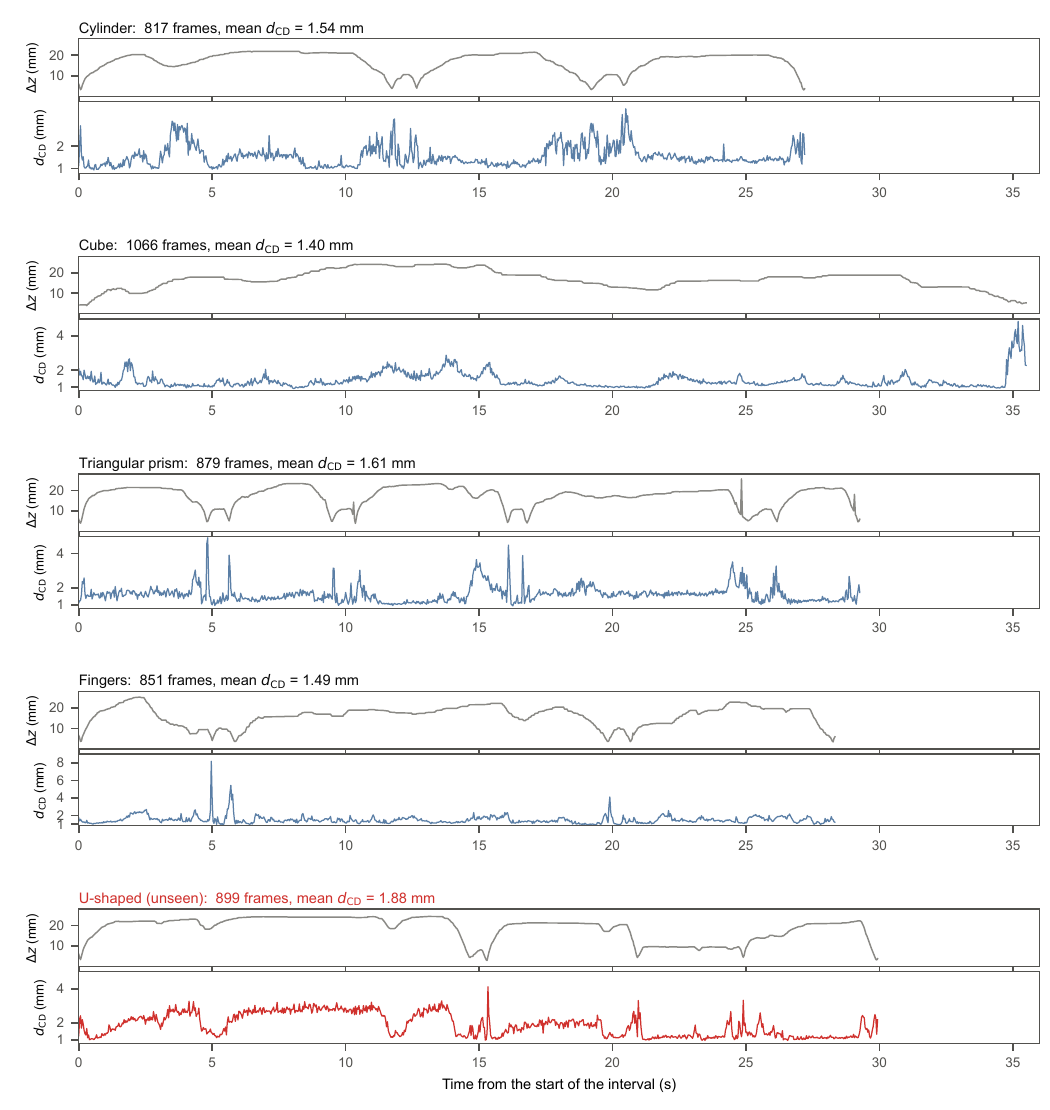}
    \caption{Deformation magnitude $\Delta z$ (gray) and reconstruction error $d_{\mathrm{CD}}$ at every frame of the five held-out intervals, in acquisition order, for the reported model. Each interval begins and ends with the membrane undeformed and contains complete loading and unloading motions. Frames within an interval are correlated.}
    \label{fig:SI-test-sequences}
\end{figure}

Figure~\ref{fig:SI-test-sequences} shows $\Delta z$ and $d_{\mathrm{CD}}$ at every frame of the five intervals.

\subsubsection{Variation between Training Runs} To estimate training variation, we trained both stages five times with the reported model configuration and random seeds 56 to 60, and evaluated each model once on the test intervals. The reported model was trained with seed 56. Table~\ref{tab:SI-seeds} shows the variation between different training seeds.

\begin{table}[!htbp]
\centering
\caption{Mean $d_{\mathrm{CD}}$ (mm) on each held-out interval for five trainings of the same configuration with different random seeds. Seed 56 is the reported model. The last column is the mean over all 4{,}512 frames.}
\label{tab:SI-seeds}
\small
\begin{adjustbox}{max width=\linewidth}
\begin{tabular}{lrrrrrr}
\toprule
Seed & Cylinder & Cube & Triangle & Fingers & U-shaped & All \\
\midrule
56 & 1.543 & 1.396 & 1.614 & 1.490 & 1.884 & 1.580 \\
57 & 1.567 & 1.404 & 1.594 & 1.519 & 1.901 & 1.591 \\
58 & 1.646 & 1.504 & 1.599 & 1.556 & 1.910 & 1.639 \\
59 & 1.681 & 1.437 & 1.508 & 1.588 & 1.734 & 1.582 \\
60 & 1.598 & 1.489 & 1.506 & 1.534 & 1.578 & 1.538 \\
\midrule
Mean $\pm$ SD & $1.607\pm0.057$ & $1.446\pm0.049$ & $1.564\pm0.053$ & $1.537\pm0.037$ & $1.801\pm0.144$ & $1.586\pm0.036$ \\
\bottomrule
\end{tabular}
\end{adjustbox}
\end{table}

\subsubsection{Contact Localization on Finger Strokes} \label{sec:SI-spatial}
\begin{figure}[!htbp]
    \centering
    \includegraphics[width=\linewidth]{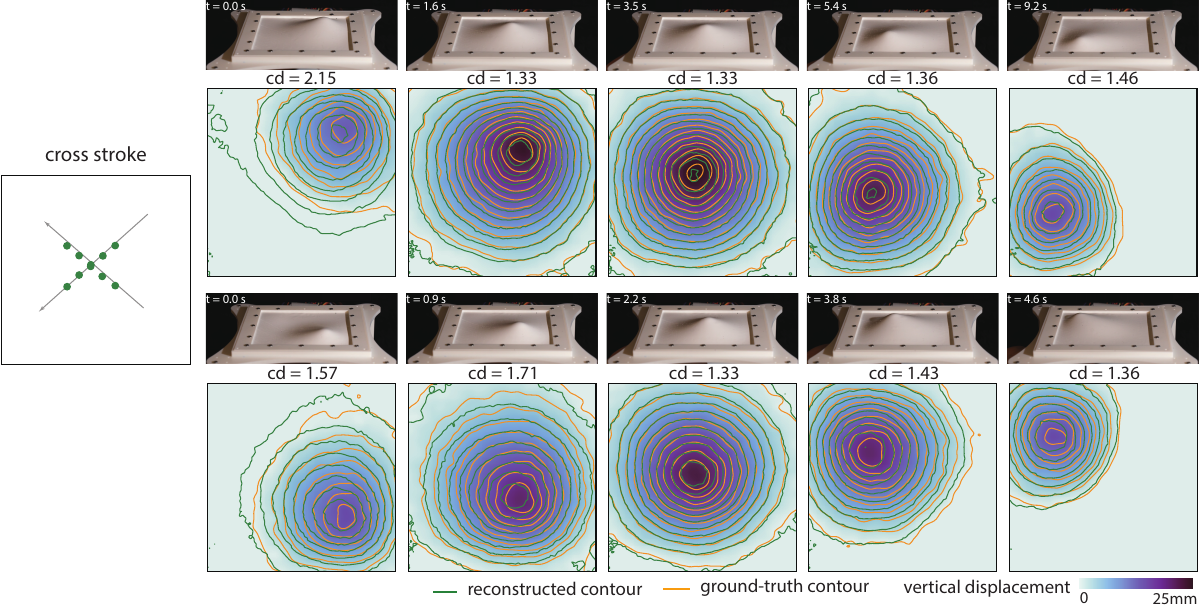}
    \caption{Reconstruction along the cross stroke, at five representative contact locations along each diagonal.}
    \label{fig:cross}
\end{figure}

\begin{figure}[!htbp]
    \centering
    \includegraphics[width=\linewidth]{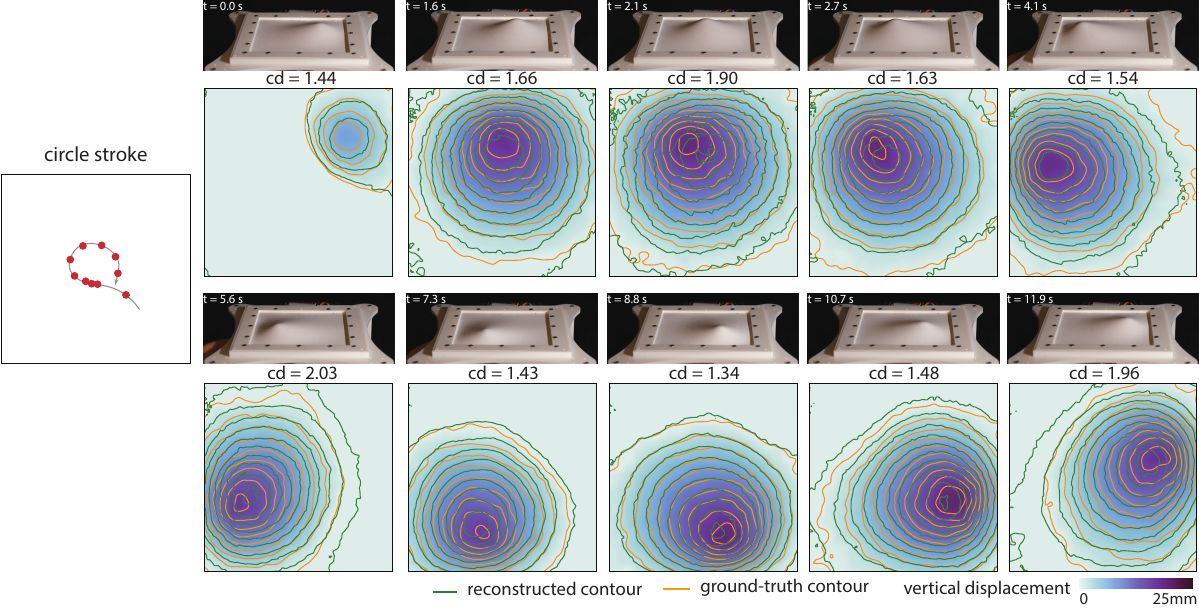}
    \caption{Reconstruction along the circle stroke, at ten representative contact locations.}
    \label{fig:circle}
\end{figure}

We additionally recorded two continuous finger strokes: a cross along both diagonals and a circle drawn near the membrane center. Figures~\ref{fig:cross} and~\ref{fig:circle} show reconstructed frames along the two strokes.

\begin{figure}[!htbp]
    \centering
    \includegraphics[width=\linewidth]{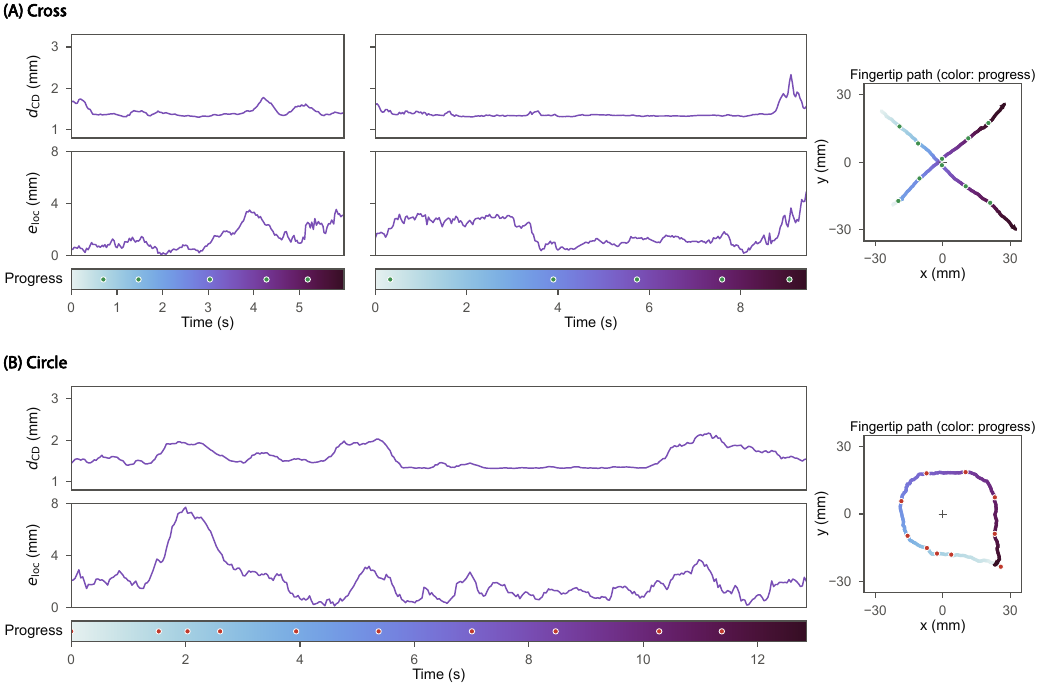}
    \caption{Reconstruction error over time on the \textbf{(A)} cross and \textbf{(B)} circle. For each stroke, the upper trace shows $d_{\mathrm{CD}}$ and the lower trace the localization error $e_{\mathrm{loc}}$. The color band below the traces and the color of the fingertip path on the right indicate only the progress along the stroke, from light at the start to dark at the end.}
    \label{fig:SI-stroke-timeseries}
\end{figure}

\begin{table}[!htbp]
\centering
\caption{Reconstruction error on the two finger strokes, over all frames of each stroke.}
\label{tab:SI-strokes}
\small
\setlength{\tabcolsep}{4pt}
\begin{adjustbox}{max width=\linewidth}
\begin{tabular}{lrrrrrrrrr}
\toprule
Stroke & Frames & \multicolumn{4}{c}{$d_{\mathrm{CD}}$ (mm)} & \multicolumn{4}{c}{Localization error (mm)} \\
\cmidrule(lr){3-6}\cmidrule(lr){7-10}
 & & Mean & Median & SD & 95th pct. & Mean & Median & SD & 95th pct. \\
\midrule
Cross & 464 & 1.407 & 1.359 & 0.128 & 1.686 & 1.607 & 1.335 & 0.943 & 3.126 \\
Circle & 386 & 1.581 & 1.522 & 0.234 & 1.975 & 1.986 & 1.694 & 1.522 & 5.997 \\
\midrule
All & 850 & 1.486 & 1.386 & 0.203 & 1.927 & 1.779 & 1.456 & 1.253 & 3.524 \\
\bottomrule
\end{tabular}
\end{adjustbox}
\end{table}

For each frame, we describe the deformation of the membrane by $d(\boldsymbol{x}) = z_0(\boldsymbol{x}) - z(\boldsymbol{x})$, the depth by which the surface at the in-plane position $\boldsymbol{x}=(x,y)$ lies below its resting height $z_0(\boldsymbol{x})$. We take the deepest point of the deformation,
\begin{equation}
\boldsymbol{c} = \operatorname*{arg\,max}_{\boldsymbol{x}} d(\boldsymbol{x}),
\label{eq:SI-contact}
\end{equation}
as the contact position. We compute the contact position $\boldsymbol{c}$ on the ground-truth surface and $\hat{\boldsymbol{c}}$ on the reconstructed surface in the same way. The localization error is the in-plane distance between the two positions,
\begin{equation}
e_{\mathrm{loc}} = \left\| \hat{\boldsymbol{c}} - \boldsymbol{c} \right\|_2.
\label{eq:SI-eloc}
\end{equation}

Table~\ref{tab:SI-strokes} and Figure~\ref{fig:SI-stroke-timeseries} summarize the results. Across both strokes, the mean localization error is 1.779~mm, with 1.607~mm on the cross and 1.986~mm on the circle. The mean $d_{\mathrm{CD}}$ is 1.486~mm, close to the 1.580~mm over the five press intervals (Table~\ref{tab:SI-test-intervals}).

\subsection{Optical Signal Repeatability}\label{sec:SI-stability}
\subsubsection{Experiment Setup}
All measurements in this section come from the 140~mm $\times$ 140~mm membrane (Figure~\ref{fig:setup}C) with 30 LEDs and 6 PDs. The tests were carried out in the following order: three sessions of repeated pressing, 500 cycles of uniaxial stretching to 30\% strain, 75 cycles to 50\% strain, and 100 cycles to 75\% strain.

\paragraph{Repeated pressing}
We pressed the membrane repeatedly by hand while the depth camera recorded its surface and the sensor recorded all 180 LED--PD readings. The recording consists of three sessions on three consecutive days (Table~\ref{tab:SI-sessions}). The sensor was switched off between days for 28.6~h and 23.0~h. S1 started about 15~min after the sensor was switched on, and S2 and S3 started at power-on. We identified pressing intervals and corresponding rest periods using the same rule defined in Section~\ref{sec:SI-indentation}, yielding 471 presses in S1 to S3 (86, 163, and 222), with a median peak of 23.9~mm and a 5th to 95th percentile range of 17.6 to 28.7~mm.

\paragraph{Repeated stretching}
After the pressing sessions, we clamped the membrane over 30~mm at two opposite edges on the tensile testing machine, leaving a gauge length of 80~mm along the direction of the flexible flat cable, so that the strain is $\Delta y/80$~mm. During setup, the membrane underwent about 450 stretch cycles at 7.5\% to 22.5\% strain. The machine then moved between $\Delta y=0$ and 24~mm (30\% strain) at 36~mm/min, and ran 500 cycles without pause over 11.2~h. Afterward, we stretched the same membrane to $\Delta y=40$~mm (50\% strain) for 75 cycles at the same speed and then to 60~mm (75\% strain) for 100 cycles. The sensor recorded all 180 channels at about 34 frames per second, and the machine recorded the displacement at 50~Hz. We matched the two clocks with a linear map fitted to the maximum-stretch times, leaving a median residual of 0.14~s.

\begin{table}[!htbp]
\centering
\caption{Repeated-pressing sessions and their reconstruction error. Each $d_{\mathrm{CD}}$ entry is the mean over the frames in the stated interval, with time measured from the start of each session. S1 started about 15~min after power-on and serves as the baseline; S2 and S3 started at power-on.}
\label{tab:SI-sessions}
\small
\begin{adjustbox}{max width=\linewidth}
\begin{tabular}{lrrrrrr}
\toprule
 & & & & \multicolumn{3}{c}{$d_{\mathrm{CD}}$ (mm)} \\
\cmidrule(lr){5-7}
Session & Duration & Frames & Presses & Whole & Warm-up & Warmed up \\
        & (min) & & & session & (0--10~min) & (10~min--end) \\
\midrule
S1 (baseline) & 14.3 & 25{,}717 & 86 & 1.187 & -- & -- \\
S2 & 36.1 & 64{,}962 & 163 & 1.217 & 1.223 & 1.214 \\
S3 & 36.2 & 65{,}076 & 222 & 1.273 & 1.356 & 1.242 \\
\bottomrule
\end{tabular}
\end{adjustbox}
\end{table}

\subsubsection{Signal Drift}\label{sec:SI-drift}

Each frame contains the readings $x_k(t)$ of the 180 LED--PD channels $k$ in 24-bit ADC counts. Drift was measured at rest, which recurs between presses. A press changes mainly the channels whose light paths cross the pressed region, so we report the drift of the channel that responds most strongly to deep presses. For each press, $t^{*}$ is the frame of maximum $d_{\max}$, and $x_k^{\mathrm{rest}}$ is the median reading over the rest period before the press. The reference $b_k^{\mathrm{start}}$ of channel $k$ is its median reading over all rest frames in the first 10~min of S1. Among the channels whose resting reading exceeds $1.28\times10^{6}$ counts, we selected the one with the largest median of $|x_k(t^{*})-x_k^{\mathrm{rest}}|/b_k^{\mathrm{start}}$ over the 161 presses with a peak deeper than 25~mm. This is L21P5. Figure~\ref{fig:characterization}A shows this channel as the relative change $\Delta I/I_0=x(t)/I_0-1$, with $I_0=b^{\mathrm{start}}$. Over 471 presses, the median lowest $\Delta I/I_0$ is $-30.5\%$.

\subsubsection{Membrane Strain during Pressing}\label{sec:SI-strain}
\begin{figure}[!htbp]
    \centering
    \includegraphics[width=0.5\linewidth]{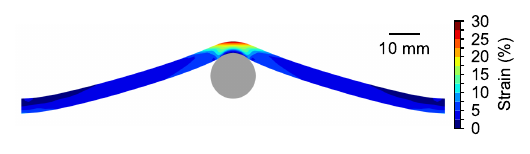}
    \caption{Simulated maximum principal engineering strain in the section through the indenter axis.}
    \label{fig:SI-fea-strain}
\end{figure}

We estimated the local strain of the membrane during pressing with a finite-element simulation in FEBio (Figure~\ref{fig:SI-fea-strain}). A 140~mm $\times$ 140~mm $\times$ 5~mm membrane clamped on all four sides was modeled as a nearly incompressible neo-Hookean solid and indented at its center by a rigid sphere of 15~mm diameter to a displacement of 20~mm, with frictionless contact. The maximum principal engineering strain reaches 28.2\% at the center of the face opposite the indenter. At the depth of the liquid-metal circuit, 0.5--1.0~mm from the indented face, it is 12.3\%.

\subsubsection{Repeatability under Cyclic Stretching}\label{sec:SI-hysteresis}

We filtered each 500-cycle test reading with a causal second-order Butterworth low-pass filter at 5~Hz, then subtracted the filtered reading of the same PD with all LEDs off. We report each channel as the relative change $\Delta I/I_0$, where $I_0$ is its reading at rest before the first cycle. For cycles 1, 250, and 500, we binned $\Delta I/I_0$ versus $\Delta y$ in 0.5~mm bins, separately for loading and unloading. The baseline of a channel is its $\Delta I/I_0$ at 0~mm on loading, its response amplitude is the change in $\Delta I/I_0$ from 0 to 24~mm on loading, and its loop width $W_r$ is the largest difference between the loading and unloading curves of the same cycle, in percentage points of $\Delta I/I_0$. We analyzed all 180 channels. Four of them, L5P1, L6P1, L26P6, and L27P6, saturated the ADC.

\begin{figure}[!p]
    \centering
    \includegraphics[width=\linewidth]{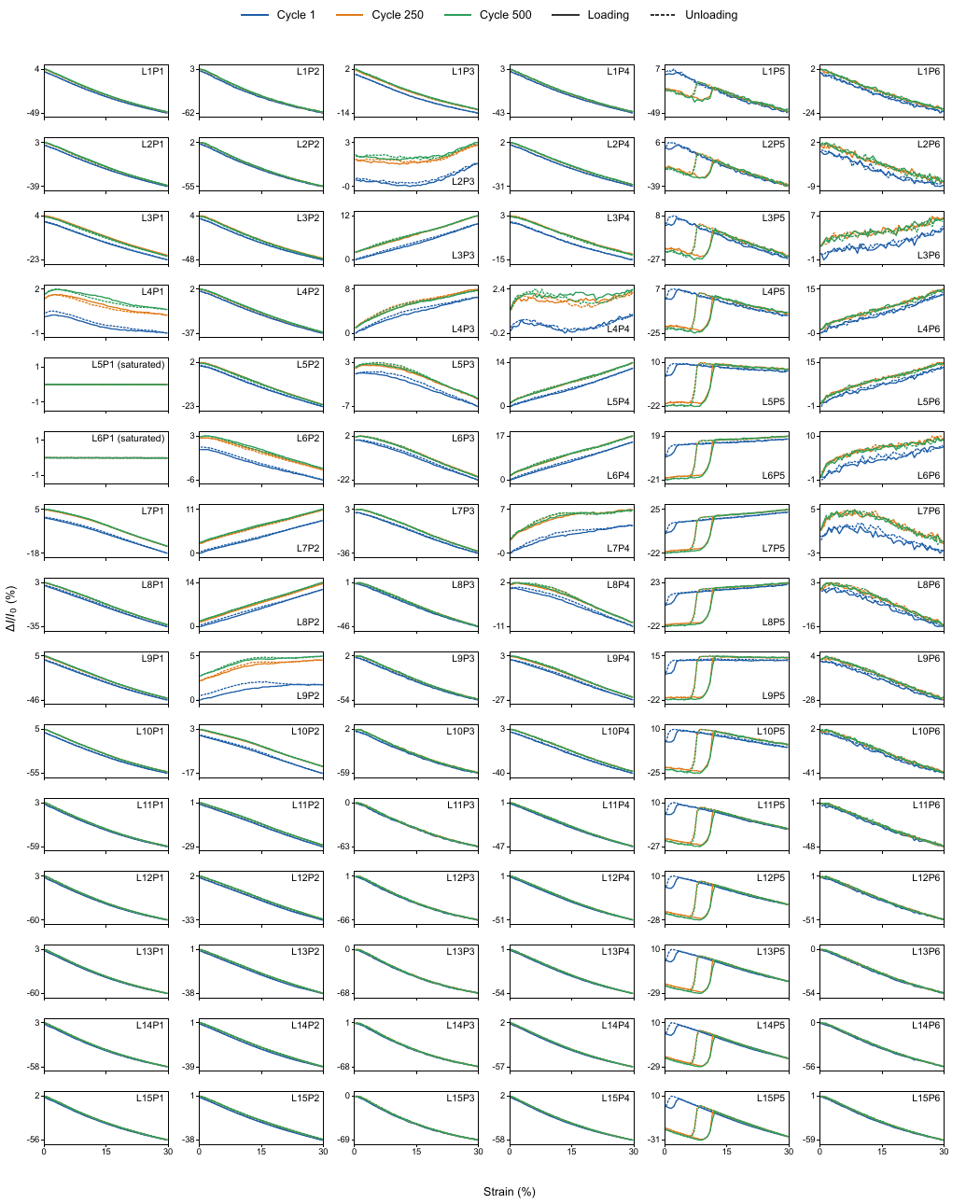}
    \caption{$\Delta I/I_0$ versus strain for channels of LEDs 1 to 15 in cycles 1, 250, and 500, up to 30\% strain, on loading (solid) and unloading (dashed). Each panel is scaled to its own range, and the numbers on the vertical axis give that range in percent.}
    \label{fig:SI-cyclic-channels}
\end{figure}

\begin{figure}[!p]
    \centering
    \includegraphics[width=\linewidth]{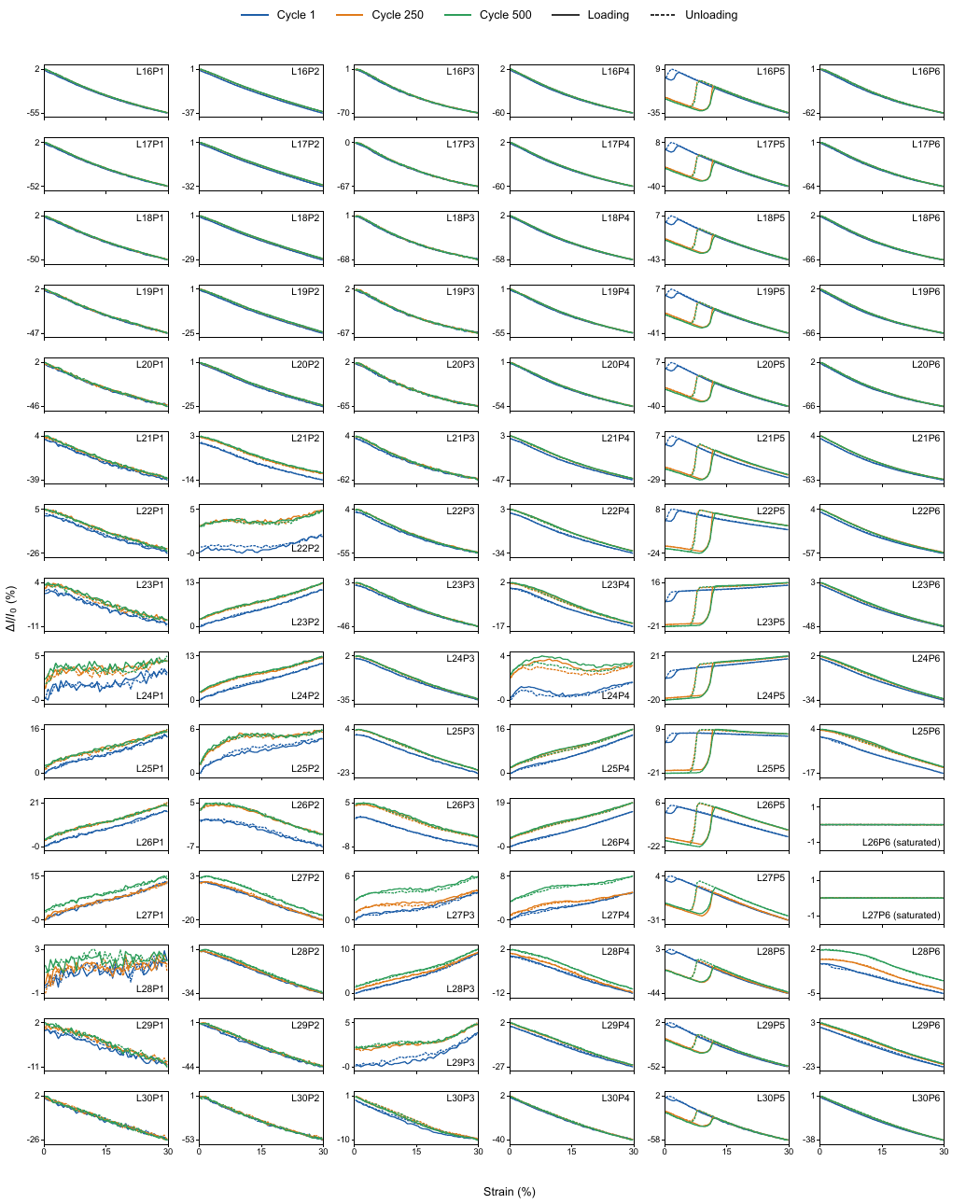}
    \caption{As Figure~\ref{fig:SI-cyclic-channels}, for the channels of LEDs 16 to 30.}
    \label{fig:SI-cyclic-channels2}
\end{figure}

Figures~\ref{fig:SI-cyclic-channels} and~\ref{fig:SI-cyclic-channels2} show the response of all 180 channels in cycles 1, 250, and 500. Over all channels, the median change from cycle 1 to cycle 500 was 1.5 percentage points for the baseline and 0.9 percentage points for the response amplitude, and the median $W_r$ was 0.9 percentage points. For comparison, a channel's reading at rest before the first cycle varied by a median of 1.4 percentage points between its 0.5th and 99.5th percentiles. Figure~\ref{fig:characterization}B,C shows the most responsive channel, L18P4, as an example.

\clearpage

\subsubsection{Extreme Stretching and Failure Modes}\label{sec:SI-stretch-failure}
At 50\% and 75\% strain, most channels' responses remained stable. The response amplitude of L15P4 was $-74.6\%$, $-73.5\%$, and $-73.4\%$ in the first, middle, and last cycles at 50\% strain, and $-85.9\%$, $-85.8\%$, and $-85.4\%$ at 75\% strain. We observed two permanent circuit failures.

\begin{figure}[!htbp]
    \centering
    \includegraphics[width=\linewidth]{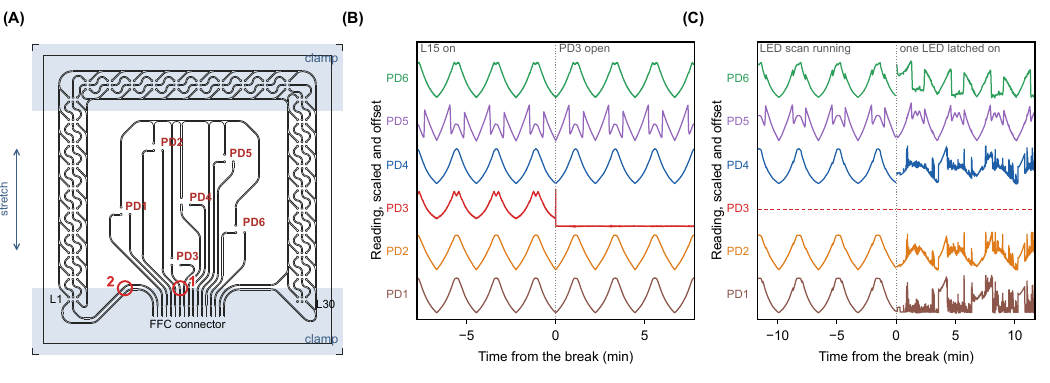}
    \caption{Failure modes under extreme stretching. \textbf{(A)} Liquid-metal trace layout with the clamped regions (shaded) and the likely sites of the PD3 open circuit (1) and the LED data line break (2). \textbf{(B)} Readings of the six PDs in the slot of L15 at 50\% strain around the open circuit of PD3. \textbf{(C)} Readings of the six PDs in the slot of L15 before and after the LED data line broke at 75\% strain. In (B) and (C), each trace is scaled to its own range and offset for clarity, and time is measured from the break. In (C), PD3 is drawn as a dashed line since it broke earlier.}
    \label{fig:SI-stretch-failure}
\end{figure}

\textbf{Open circuit of PD3 after 60 cycles at 50\% strain.}
In cycle 61 at 50\% strain, all readings of PD3, including the reading with all LEDs off, jumped to the same constant value independent of the light (Figure~\ref{fig:SI-stretch-failure}B). This is an open circuit in the PD3 wiring. At 75\% strain, PD3 briefly reconnected near zero strain in most cycles, suggesting a crack in a trace. PD3 is the PD closest to the bottom clamp, and its traces cross the clamp edge at site 1 in Figure~\ref{fig:SI-stretch-failure}A, where the strain concentrates. The other five PDs kept working through the rest of the test.

\textbf{LED data line breaks after 90 cycles at 75\% strain.}
In cycle 91 at 75\% strain, near $\Delta y=56$~mm, the LED scan stopped, and every slot afterward recorded the light of one LED that stayed on. The LED chain therefore lost its data signal upstream of the first LED, most likely where the LED bus crosses the edge of the bottom clamp on its way from the connector to L1 (site 2 in Figure~\ref{fig:SI-stretch-failure}A). The liquid-metal wiring of the PDs and the LED power lines kept working, and all PDs except PD3, whose wiring had already failed at 50\% strain, continued to follow the stretching cycles under the light of this LED until the end of the 100 cycles (Figure~\ref{fig:SI-stretch-failure}C).

Future work could improve the robustness of the liquid-metal circuit under large repeated strain, for example by using serpentine interconnects.

\subsection{Feature Importance Analysis}\label{sec:SI-optimization}
On an older membrane prototype with 30 LEDs and 5 PDs, we further conducted an ablation study to understand the contributions of individual PDs and LEDs to the reconstruction network and to guide future sensor iterations. The number of LEDs and photodiodes embedded within the waveguide is critical for obtaining a \textit{sufficient} set of independent measurements of light intensity to reconstruct the global geometry. However, increasing the number of components also increases the fabrication complexity and limits data readout speed. Therefore, identifying the minimum number of LEDs and PDs that provide sufficient information for accurate shape reconstruction is essential for efficient membrane-sensor design.

\paragraph{Feature Importance via SAGE}
To quantitatively evaluate each optoelectronic component's contribution to reconstruction performance, we adopt the Shapley additive global importance (SAGE) value \cite{covert2020understanding, covert2021explaining}, an \textit{additive}, model-agnostic metric that attributes predictive power to individual input features (or groups of features). We compute the importance of each LED and PD in the trained PD-to-latent MLP $h(\cdot)$ using the group SAGE value evaluated over predefined feature sets. Specifically, the $i$-th LED group is defined as $\mathcal{G}_{L}^{(i)}=\{\ell(j-1)+i\mid j=1, \dots,p\}$, corresponding to the $p$ photodiode readings collected when LED $i$ is activated. Similarly, the $j$-th PD group is defined as $\mathcal{G}_{P}^{(j)}=\{\ell(j-1)+i\mid i=1,\dots,\ell\}$, which represents the reading of PD $j$ across all $\ell$ LEDs. These groupings correspond physically to adding or removing a single LED or PD from the system. Let $\boldsymbol{V}=[V_1, \dots, V_{p\ell}]^{\top}$ denote the random vector of input features and let $\boldsymbol{Z}\triangleq E(S_{\mathrm{gt}})$ denote the random vector of the latent code obtained by encoding the ground-truth surface $S_{\mathrm{gt}}$. We define the index set of all features as $\mathcal{N}=\{1,\dots,p\ell\}$. For any subset $\mathcal{S}\subseteq\mathcal{N}$, the corresponding subset of features is denoted by $\boldsymbol{V}_{\mathcal{S}}\triangleq\{V_{i}|i\in\mathcal{S}\}$. The predictive power of a feature subset indexed by $\mathcal{S}$ is defined in Equation~(\ref{eq3})
\begin{equation}
    u\left(\mathcal{S}\right)=\mathbb{E}\left[\mathcal{L}_{\mathrm{mse}}\left(h_{\varnothing}\left(v_{\varnothing}\right),Z\right)\right] - \mathbb{E}\left[\mathcal{L}_{\mathrm{mse}}\left(h_{\mathcal{S}}\left(v_{\mathcal{S}}\right),Z\right)\right]
    \label{eq3}
\end{equation}
where $h_{\mathcal{S}}\left(v_{\mathcal{S}}\right)=\mathbb{E}\left[h(V)|V_{\mathcal{S}}=v_{\mathcal{S}}\right]$ denotes the reduced model obtained by marginalizing the excluded features $V_{\mathcal{N}\backslash\mathcal{S}}$. The SAGE value associated with a specific LED or PD group $\mathcal{G}$ is then defined by Equation~(\ref{eq4}).
\begin{equation}
    \Phi_{\mathcal{G}}=\frac{1}{|\mathcal{N}|}\sum_{\mathcal{S}\subseteq \mathcal{N}\backslash\mathcal{G}}\binom{|\mathcal{N}|-1}{|\mathcal{S}|}^{-1}\left[u(\mathcal{S}\cup\mathcal{G})-u(\mathcal{S})\right]
    \label{eq4}
\end{equation}

\begin{figure}[!htbp]
    \centering
    \includegraphics{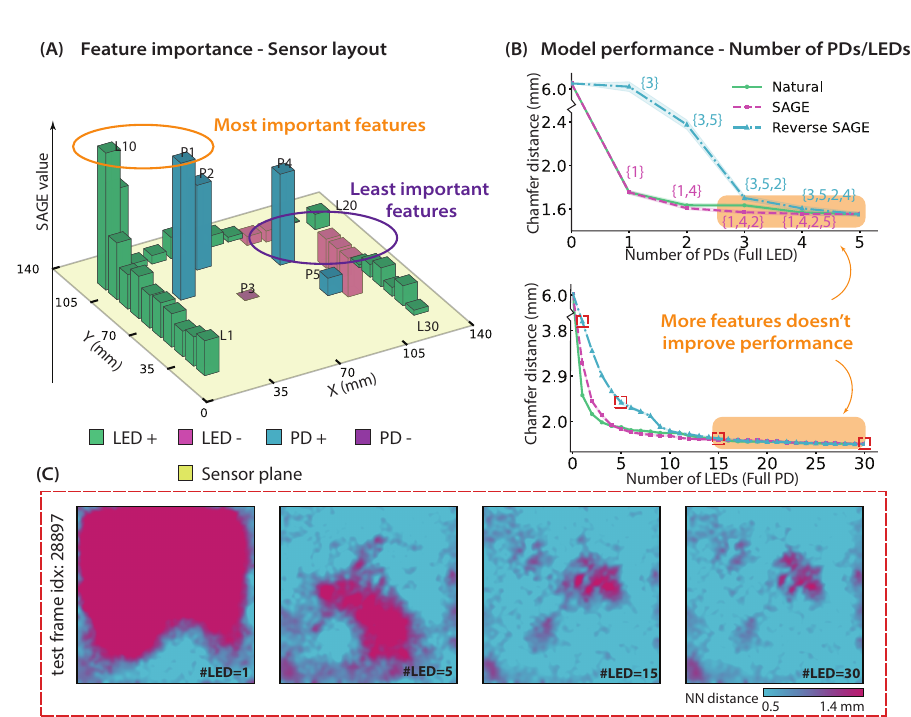}
    \caption{\textbf{Feature importance analysis of the shape reconstruction model.} \textbf{(A)} Single feature (LED/PD) importance evaluation based on SAGE values and the relation with physical locations. \textbf{(B)} Reconstruction performance under progressive feature inclusion. Top: PDs are added one by one with all LEDs enabled; Bottom: LEDs are added one by one with all PDs enabled. Curves compare different inclusion orders (Natural, SAGE-ranked, and reverse SAGE). \textbf{(C)} Sample reconstruction error shown as a heatmap of NN distance of representative reduced models with different numbers of LEDs.}
    \label{fig:optimization}
\end{figure}

\paragraph{Guidance on Sensor Layout}
Using the group SAGE value, we quantify the importance of each LED and PD and relate these scores to their physical locations within the sensor (Figure~\ref{fig:optimization}A). Notably, PD group $\mathcal{G}_{P}^{(3)}$, which is located near the center of the membrane, exhibits the lowest importance value, and the LED groups $\mathcal{G}_{L}^{(15)}$ to $\mathcal{G}_{L}^{(30)}$ generally exhibit lower importance. We hypothesize that these reduced importance scores arise from the baseline light intensity measured by the photodiode: when the PD's baseline intensity is low, either due to a large LED--PD distance or voltage drops along the LED power traces, small deformation-induced changes become harder to resolve relative to noise, thereby reducing the informativeness of those channels. This observation suggests that PD placement can be optimized to maintain baseline intensity within a favorable operating range, ensuring sufficient sensitivity for each measurement channel.

We further leverage this importance analysis to study how model performance scales with the number of optoelectronic components via a progressive feature-inclusion experiment. Specifically, we select $K$ PD or LED groups $\mathcal{G}^{(i)}$ to form a reduced feature index set $\mathcal{S}=\bigcup_{i=1}^{K} \mathcal{G}^{(i)}$, retrain the reduced MLP $h_{\mathcal{S}}$, and evaluate the reconstruction error of the composed model $f_{\mathcal{S}}=D\circ h_{\mathcal{S}}$ using the average Chamfer distance on the test set. Feature groups are added in three orders: SAGE (most informative first), reverse SAGE (least informative first), and natural (numerical order). The results presented in Figure~\ref{fig:optimization}B show that importance-aware inclusion yields a clear advantage at small $K$ values: selecting the most informative groups first generally achieves lower error than including the least informative groups. However, the marginal benefit of adding more optoelectronic components decreases as indicated by the plateau in the curve. The nearest-neighbor (NN) distance error maps shown in Figure~\ref{fig:optimization}C also indicate that the regions of large reconstruction error shrink significantly as the most informative channels are incorporated, with diminishing improvements for large values of $K$. Together, these results suggest that the optical signal becomes redundant after adding a moderate number of channels (i.e., 15 LEDs), providing a practical upper bound for future design. The LED count can be reduced substantially with minimal loss in accuracy, while PDs should be placed strategically to maintain high information-density measurements.

\clearpage
\phantomsection
\addcontentsline{toc}{section}{References}
\setlength{\bibsep}{3pt}
\bibliography{Bibtex/reference}
\end{document}